# Artificial Persuasion in Pedagogical Games

[A Book Draft]

by

Zhiwei Zeng

23 Jan 2016

# Table of Contents

# List of Figures

List of Tables    V

# List of Tables

# Abstract

A Persuasive Teachable Agent (PTA) is a special type of Teachable Agent which incorporates a persuasion theory in order to provide persuasive and more personalized feedback to the student. By employing the persuasion techniques, the PTA seeks to maintain the student in a high motivation and high ability state in which he or she has higher cognitive ability and his or her changes in attitudes are more persistent. However, the existing model of the PTA still has a few limitations. Firstly, the existing PTA model focuses on modelling the PTA's ability to persuade, while does not model its ability to be taught by the student and to practice the knowledge it has learnt. Secondly, the quantitative model for computational processes in the PTA has low reusability. Thirdly, there is still a gap between theoretical models and practical implementation of the PTA.

To address these three limitations, this project proposes an improved agent model which follows a goal-oriented approach and models the PTA in its totality by integrating the Persuasion Reasoning of the PTA with the Teachability Reasoning and the Practicability Reasoning. The project also proposes a more abstract and generalized quantitative model for the computations in the PTA. With higher level of abstraction, the reusability of the quantitative model is also improved. New system architecture is introduced to bridge the gap between theoretical models and implementation of the PTA.

An instance of the PTA is also implemented and embedded into a 3D pedagogical game called VS Saga to demonstrate the practice of instantiating a PTA from the proposed agent model and system architecture. A focus group study has been conduct to assess the PTA qualitatively. The results of the focus group study are quite positive, showing the PTA has the ability to motivate and engage the student by generating personalized feedback and the potential to help the student develop positive attitudes towards learning. Positive feedback has also been received for the completeness of the new agent model and the reusability of the proposed system architecture. As for future studies, quantitative assessment of the PTA still needs to be carried out in order to analyze the effectiveness of the PTA statistically. The PTA also needs to build up its flexibility to cater to different learning topics.

# Chapter 1 Introduction to Persuasive Teachable Agent

## 1.1   Motivation

Teachable Agents (TAs) are educational agents which have emerged from the interdisciplinary research in education, computer science and psychology. The design and development of a Teachable Agent is enlightened by the Learning-by-Teaching pedagogy which entails learning by teaching others [1]. The TA plays the role of a tutee which requires tutoring from the student. During the process of teaching the agent tutee, the student can enhance his or her comprehension and reflection of content knowledge, acquire inquiry skills and stay motivated due to his or her sense of responsibility towards the agent tutee [2].

In order to harness the benefits of Learning-by-Teaching, researchers have designed various TAs, such as Betty's Brain [3] and SimStudent [4]. However, these TAs generally lack of the ability to interact with the student spontaneously and maintain the student's attention. They are also generally incapable of giving personalized responses or feedback to their student tutors. More recently, attempts have been made to improve on these drawbacks. Affective Teachable Agent improves TA's spontaneity and believability by incorporating goal orientation feature and generating emotional responses according to the student's teaching [5]. Dynalearn is another project which improves TA's ability to generate feedback and recommendations by making use of a qualitative reasoning model [6]. Most of the aforementioned TAs would provide feedback to the student based on their learning outcome. There are two limitations with learning with these TAs: (1) the student may not have mastered the topic himself or herself when teaching; (2) the student is not fully motivated to teach the TA.

A Persuasive Teachable Agent (PTA) is a special TA which has the ability to be taught, to practice knowledge learnt and to persuade the student to teach it [20]. Taking a different approach from the aforementioned attempts, a PTA seeks to induce long lasting attitude changes of the student towards learning by incorporating a persuasion theory [2]. It also seeks to improve TA's responsiveness and the quality of feedback generated. According to the PTA model that Lim has proposed in her study [20], the motivation and ability level of the student is tracked and measured constantly while he or she is interacting with and teaching the PTA. The PTA generates appropriate and personalized persuasion cues based on the student's measured motivation and ability level, in order to maintain the student in a state in which he or she has higher cognitive ability and his or her changes in attitudes are more persistent [2].

The PTA addresses several issues of previous TAs, including lack of spontaneity and personalized feedback. However, it still has **a few limitations**. Firstly, the PTA model focuses on modelling PTA's ability to persuade, while does not model its ability to be taught and to practice. Secondly, the quantitative model for computational processes in the PTA is difficult to reuse. Thirdly, there is still a gap between theoretical models and practical implementation of PTA. This project looks into these limitations, and aims to address them.

With the rise of consumer games on various devices, including PCs, smartphones and tablets, pedagogical games are also gaining popularity. Researches have suggested that videogames could become powerful learning tools which could be adapted to incorporate different curriculums [7]. Videogames create immersive learning environments with appealing video and audio effects and most of them are created in 2D or 3D environments. Studies on pedagogical videogames indicate that they can help the student master content knowledge [8] and acquire practical skills [9]. Besides, the virtual environments in videogames provide a digital learning ambience in which various types of student behavioral data can be tracked and recorded.

Teachable Agents can also be embedded into pedagogical videogames [10]. TAs appear in the gaming environment as avatars which are capable of interacting with students and performing actions [11]. Based on the student's behavioral data collected in virtual gaming environments, videogames enable the embedded TAs to provide more immediate and interactive feedback to the student. More specifically, in terms of Persuasive Teachable Agents, collected data can also be used to direct the selection of appropriate persuasion strategies.

As mentioned before, the existing PTA model still has a few limitations. This project **aims to address the limitations** by:

(1) Redefining a PTA agent model that depicts PTA's ability to persuade, integrated with its ability to learn and to practice

(2) Proposing an improved quantitative model for computational processes in PTA

(3) Introducing system architecture to guide PTA implementation.

Considering the favorable effects that educational games and virtual environments have on learning, an instance of the PTA is also implemented and embedded into a pedagogical game in 3D virtual environment. This instance can be used subsequently for the assessment and evaluation of the PTA.

## 1.2   Research Objectives

By incorporating the idea of persuasion, the PTA can achieve greater spontaneity, generate more personalized feedback and induce long lasting attitude changes towards learning. However, the **existing PTA model has several limitations**:

(1) *Persuasion Reasoning of PTA is not integrated with Teachability and Practicability Reasoning.* A PTA should be able to be taught, to practice and to persuade. The existing model focuses on depicting the PTA's ability to persuade, but does not describe the PTA's ability to be taught and to practice.

(2) *The quantitative model in Persuasion Reasoning of PTA is difficult to reuse.* Existing PTA model employs a fuzzy tool, Fuzzy Cognitive Map (FCM), to model the computational processes in Persuasion Reasoning. However, the quantitative FCM model is highly context dependent thus difficult to be reused in a different context.

(3) *Gap between theoretical models and practical implementation.* The agent model has a high level of abstraction. It is hard to directly instantiate a PTA from the existing PTA model. Thus, another layer of design is needed between the agent model and implementation, which can provide more detailed guidelines for the implementation of PTA.

This project aims to address the limitations mentioned above, and the objectives of the project are list as below:

- To redefine a complete PTA model by integrating the Persuasion Reasoning, the Teachability Reasoning and the Practicability Reasoning. The new agent model should describe the PTA's ability to be taught and to practice, in addition to its ability to persuade.
- To propose a context independent quantitative model for the computational processes in the Persuasion Reasoning of the PTA, by increasing the level of abstraction and generalizing from the original model.
- To propose system architecture to provide more guidelines for implementation. The proposed system architecture should be detailed enough, so that a PTA can be easily instantiated from it. However, at the same time, it also needs to retain certain level of abstraction, so that it does not restrict the way of implementation.
- To implement an instance of the PTA. This instance can demonstrate the practice of instantiating a PTA from the proposed agent model and system architecture.
- To assess the PTA. As an instance of the PTA has been developed, it can be used for the assessment and evaluation of the PTA.

## 1.3  Contributions

The main contributions of this project include:

- *A complete and integrated PTA agent model.* The agent model depicts a PTA **in its totality**, which includes:
    - *Persuasion Reasoning.* The persuasion reasoning model in existing PTA model is simplified and integrated with the Teachability Reasoning and the Practicability Reasoning.
    - *Teachability Reasoning.* A teachability reasoning model is added to depict the PTA's ability to be taught.
    - *Practicability Reasoning*: A practicability reasoning model is added to depict the PTA's ability to practice.
- *An improved quantitative model for Persuasion Reasoning of the PTA.* The quantitative model can be reused in different contexts. For example, it can be reused for PTAs in different knowledge domains and with different implementations.
- *New system architecture for the PTA.* The system architecture mainly describes how a PTA can be implemented, especially in terms of control structures. The design can be reused for multiple PTAs.
- *Deploying a PTA in pedagogical game in 3D virtual environment.* An instance of PTA has been implemented in a 3D virtual educational game. The implementation has followed the improved agent model and proposed system architecture and demonstrated all major characteristics of a PTA.
- *A focus group study to assess the PTA.* A focus group study including a group of 6 participants has been conducted to assess the PTA qualitatively. In order to collect feedback and comments on various aspects of the PTA, the focus group is consisted of secondary school teachers, agent researcher and game developers.

## 1.4  Report Organization

The chapters of this report are organized as follow:

Chapter 2 reviews related researches, studies and experiments in the area of Teachable Agent and Persuasive Teachable Agent. As the implemented PTA is embedded within a game in 3D environment, related work on educational games and virtual environments are also presented.

Chapter 3 redefines PTA and introduces the improved agent model.

Chapter 4 introduces the proposed system architecture.

Chapter 5 demonstrates the practice of implementing a PTA following the proposed agent model and system architecture.

Chapter 6 presents the case studies of the implemented PTA. It also describes the approach to assess the PTA and summarizes the assessment results.

Chapter 7 concludes this project and discusses possible areas for future work.

# Chapter 2 Literature Review

## 2.1 Teachable Agents

The idea of Teachable Agent (TA) is inspired by the Learning-by-Teaching pedagogy which entails learning by teaching others [1]. During a peer tutoring program, Gaustard found that student tutors usually benefited more or at least the same as their peer tutees [12]. With the development of artificial intelligence, the TA is developed to take the role of a tutee and can be taught by students.

Generally, the TA facilitates and benefits learning process from the following three perspectives [11]: (1) helping students structure and organize their knowledge; (2) invoking students' sense of responsibility towards teaching and their own learning; and (3) enhancing students' knowledge reflection and metacognition.

During the recent two decades, TA has been an active research area. Researchers have developed various TAs for different purposes and curriculums in various formats.

Betty's Brain is an agent developed to facilitate the learning of middle school natural science topics [3]. The student can teach Betty by completing concept maps. Concept maps help the student to structure and reorganize his or her knowledge while teaching. The student is able to assess Betty's learning by querying Betty or put her through a quiz and observe the answers or the quiz outcome.

SimStudent is another TA which simulates a classroom learner and can be taught in the domain of mathematics [4]. SimStudent is able to learn cognitive skills from the student by analyzing his or her inputs. The student tutors the agent by solving an algebraic equation collaboratively with the agent in a stepwise fashion. A test can be used to assess SimStudent's learning performance after tutoring.

However, both Betty's Brain and SimStudent interact with students passively and are incapable of giving personalized feedback. Affective Teachable Agent is introduced to improve the spontaneity and believability of the TA [5]. Affective Teachable Agent has incorporated goal-orientation feature. While pursuing its predefined goals, the agent is able to take spontaneous actions to interact with the student. The agent has also employed affective computing by simulating emotions to build an emotional tie with the student.

DynaLearn is an agent designed to facilitate the learning of conceptual knowledge in science subjects. The student teaches the agent by creating conceptual models in a graphical editor. The agent goes through qualitative reasoning to generate feedback and recommendations to the student. Similar to Affective Teachable Agent, DynaLearn also simulates simple emotions to bind the student and the agent.

Each of the four aforementioned attempts has made their own innovations to improve learning experiences. However, they also have their own drawbacks. A summary of these TAs are presented in Table 2.1 below.

Table 2.1 Summary of Teachable Agents [20]

| Teachable Agent | Issues | Innovations | Application Domain |
| --- | --- | --- | --- |
| Betty's Brain | • Interact with students passively<br>• Incapable of generating personalized feedback<br>• Students not fully engaged | Artificial Intelligence | Science |
| SimStudent | | Machine Learning | Mathematics |
| Affective Teachable Agent | • Students not fully engaged in learning with TA | Goal Orientation Affective Computing | Lower Secondary Science |
| DynaLearn | • Lack of clarity and relevance of feedback<br>• Students not fully engaged | Qualitative Reasoning | Science Concepts |

Even for the improved TA models such as Affective Teachable Agent and DynaLearn, there are still **two issues with learning with TAs** [20]:

(1) *Students may not have mastered the topic themselves when teaching the TA*. This may cause the student to commit errors while teaching. In this situation, it is hard to give constructive feedback by solely referring to the TA's learning performance. It is crucial to understand the student's ability, i.e. how much he or she knows by the time he or she teaches, in order to give adequate feedback.

(2) *Students are not fully engaged in teaching the TA*. In order to harness full benefits of Learning-by-Teaching, the student needs to actively engage in teaching. However, it is hard to keep the student motivated all the time, especially when the TA is passive in interactions and incapable of giving personalized responses. Thus, it is important to monitor the student's motivation level and generate corresponding feedback.

To address these two issues that most existing TAs have, an improved TA model should have the ability to provide responses and feedback based on the student's ability and motivation, which ensures that the student possesses necessary ability required for teaching the topic and is constantly motivated to teach.

## 2.2 Persuasive Teachable Agents

A Persuasive Teachable Agent (PTA) is a special TA which has the ability to be taught, to practice and to persuade the student to teach it. Persuasion is a kind of interaction between human beings, which aims to influence others by changing their attitudes [13]. Originated from the psychology and the communications domain, persuasion theories have increasing applications in the human computer interaction domain, in the attempt to induce attitude changes of human beings. The PTA seeks to induce long lasting attitude changes of students towards learning by incorporating a persuasion theory [2].

The persuasion theory that has enlightened the idea of the PTA is the Elaboration Likelihood Model (ELM). The ELM is a dualistic model, which states that there are two routes of persuasion, namely the central route and the peripheral route [14]. According to the ELM, one is more willing to go through complicated cognitive processes if he or she is motivated and possesses the necessary abilities required to perform a behavior. After deep cognitive processes, one is likely to go through the central route of persuasion where his or her attitude changes tend to be enduring. On the contrary, one would avoid complicated cognitive processes if he or she is not motivated or does not possess the necessary abilities. Without any deep consideration, one is likely to go through the peripheral route of persuasion where the resulting attitude changes tend to be less enduring.

PTA has two **major characteristics**:

> (1) *Goal-orientation*. Similar to Affective Teachable Agent [5], the PTA is goal-oriented. The primary goal of the PTA is to keep students on the central route of persuasion, so that their positive attitude changes towards learning and the knowledge they obtained could be more enduring. The PTA acts towards its goal by persuasion.
>
> (2) *Ability to persuade*. The student's motivation and ability level are tracked and measured constantly while he or she is interacting with and teaching the PTA. According the ELM, when the student is not motivate or does not possess the abilities required to teach the PTA, he or she is likely to go through the peripheral route of persuasion. In this situation, the PTA generates appropriate and personalized persuasion cues to direct the student back on to the central route.

With its goal-orientation and persuasion features, the PTA is able to provide immediate and personalized feedback based on the student's ability and motivation, which previous TAs are incapable of.

In her study, Lim has propose a PTA model [2], as shown in Figure 2.1. The PTA receives environment data from the Event Tracker and performs onto its environment by carrying out actions. The Knowledge Base stores the knowledge the PTA has obtained from its student tutor.

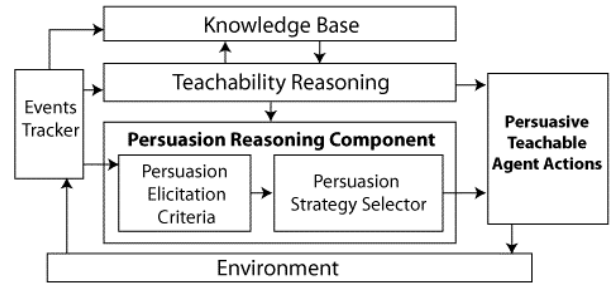

Figure 2.1 Existing PTA Model [2]

The Teachability Reasoning has two cycles, the learning cycle and the reasoning cycle. In this two cycles, the PTA learns from the student tutor and responds to his or her inquiries respectively.

The Persuasion Reasoning determines the student tutor's motivation and ability level according to the ELM and Fuzzy Cognitive Map (FCM) [15]. The Persuasion Reasoning is modelled based on Goal Net [16], a methodology for modelling goal-oriented agents.

However, as mentioned before, the **existing PTA model still has a few limitations**. After reviewing the existing PTA model, these limitations can be described more specifically as below:

(1) *Persuasion Reasoning of PTA is not integrated with Teachability and Practicability Reasoning.* The existing agent model has only proposed the Goal Net model for the Persuasion Reasoning while does not have Goal Net models for the Teachability Reasoning and the Practicability Reasoning.

(2) *The quantitative model in Persuasion Reasoning of PTA is difficult to reuse.* Existing PTA model employs a fuzzy tool, Fuzzy Cognitive Map (FCM), to model the computational processes in Persuasion Reasoning. However, the quantitative FCM model is highly context dependent thus difficult to be reused in a different context.

(3) *Gap between theoretical models and practical implementation.* The agent model has a high level of abstraction. It is hard to directly instantiate a PTA from the existing PTA model. Thus, another layer of design is needed between the agent model and implementation, which can provide more detailed guidelines for the implementation of PTA.

To address these three limitations, firstly, a complete agent model has to be proposed. A complete agent model should include not only the Goal Net modelling for the Persuasion Reasoning, but also Goal Net modellings for the Teachability and Practicability Reasoning. Secondly, a more abstract and generalized FCM model is required for the quantitative processes in the PTA. Thirdly, the gap between theoretical models and implementation of the PTA should also be bridged. In order to assess and evaluate the PTA, an instance of the PTA also needs to be implemented.

## 2.3 Educational Games and Virtual Environments

Nowadays, games are flourishing on various devices. As digital natives, most of the students today have played games and some even made game an integral part of their lives. Games, especially video games, are highly engaging and are good at content delivering due to their appealing audio and video effects. Studies on pedagogical videogames indicate that they can help students master content knowledge [8] and acquire practical skills [9].

Most videogames are deployed in 2D or 3D environments. River City is a multi-user 3D virtual environment built by Chris Dede's group at Harvard University to facilitate learning in the science domain [17]. After two weeks, the group of students who played in River City improved their content knowledge and inquiry skills by a greater extent compared to the control group which followed the normal paper-based curriculum.

Teachable Agents can also be embedded into pedagogical videogames [10]. TAs appear in the gaming environment as avatars which are capable of interacting with students and performing actions [11]. Affective Teachable Agent has been deployed in the 3D virtual educational game, Virtual Singapura. Based on the student's behavioral data collected in virtual gaming environment, Affective Teachable Agent is capable of providing appropriate emotional responses to the student. The assessment Affective Teachable Agent has proved its effectiveness not only in improving students' learning outcome, but also in motivating students and fostering their self-efficacy [11].

To harness the benefits of educational games and virtual environments, an instance of the PTA is implemented and embedded into a 3D pedagogical game called VS Saga. The implementation of this instance demonstrates the practice of instantiating a PTA from the proposed agent model and system architecture. It can also be used subsequently for the assessment and evaluation of the PTA.

# Chapter 3 Design of Persuasive Teachable Agent

## 3.1 Redefining PTA

**Definition 1**: As illustrated in Figure 3.1, a Persuasive Teachable Agent is an agent which can be defined by a tuple **PTA = (E, Et, K, Rs, R, A)**, where

**E** is the set of environmental states that can be perceived by the agent;

**Et** is the set of percepts/ events that the agent perceives from the its environment;

**K** is the set of knowledge that agent maintains;

**Rs** is the set of selection mechanisms which the agent adopts to select a reasoning in **R** to go through in each cycle;

**R** is the set of reasoning the agent is able to perform, **R** can be further defined by the tuple **R = (P, Tr, Pr)**;

**A** is the set of actions that the agent is able to perform onto the environment.

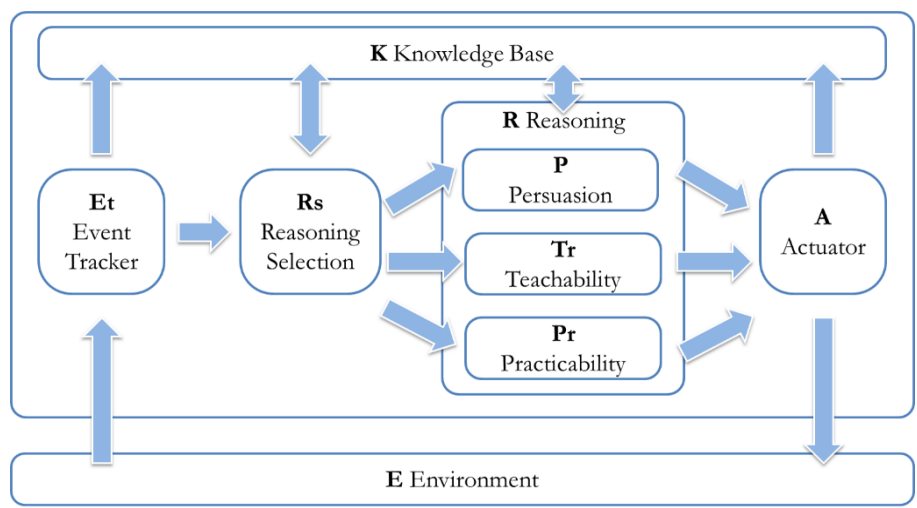

Figure 3.1 Improved PTA Model

**Definition 2**: **R** can be defined by the tuple **R = (P, Tr, Pr)**, where **P** stands for **Persuasion**, which is the reasoning mechanism that enables the agent to persuade the student.

**Tr** stands for **Teachability Reasoning**, which is the reasoning mechanism that enables the agent to learn from its student tutor.

**Pr** stands for **Practicability Reasoning**, which is the reasoning mechanism that enables the agent to practice the knowledge learnt.

**Definition 3**: According the proposed new PTA model, the agent can go through 3 different cycles:

Persuasion Cycle: EtRsP(KA)

(1) **Perceive**: The agent perceives environmental states/ events.

(2) **Reasoning Selection**: The agent selects an appropriate reasoning according to its percepts.

(3) **Persuasion Reasoning**: The agent goes through the reasoning for persuasion. It determines whether persuasion is required by evaluating the student's current motivation and ability level. If persuasion is required, it selects appropriate persuasion cue according to its percepts.

(4) (**Knowledge Base**): If persuasion is required, the agent retrieves the selected persuasion cue from the knowledge base.

(5) (**Action**): If persuasion is retrieved, the agent executes the cue.

Teaching Cycle: EtRsTrK

(1) **Perceive**: The agent perceives environmental states/ events.

(2) **Reasoning Selection**: The agent selects an appropriate reasoning according to its percepts.

(3) **Teachability Reasoning**: The agent goes through the reasoning for teachability. It acquires knowledge from the student tutor. It comprehends and translates the knowledge into knowledge representations that can be accepted by the knowledge base.

(4) **Knowledge Base**: The agent stores the knowledge acquired into the knowledge base.

Practicing Cycle: EtRsPr(K)A

(1) **Perceive**: The agent perceives environmental states/ events.

(2) **Reasoning Selection**: The agent selects an appropriate reasoning according to its percepts.

(3) **Practicability Reasoning**: The agent goes through the reasoning for practicability. It determines how to respond to the query of its student tutor according to the knowledge it has acquired. If responses of the agent are pre-determined, the agent chooses the correct response.

(4) (**Knowledge Base**): If responses are pre-determined, the agent retrieves the selected response from the knowledge base.

(5) **Action**: The agent responses to the query of its student tutor.

**Definition 4**: Knowledge **K** of the PTA consists of the following five subsets:

| | |
|---|---|
| **Goal Net Model** | is a set of Goal Net structures that defines the agent's goal-orientation characteristics and drives it during its goal pursuit. (Goal Net Model with be described in section 4.1.2 ~ 4.1.5.) |
| **FCM Model** | stores the FCM structure used by the agent to derive the values of motivation and ability from their constituent factors. |
| **Domain Knowledge** | refers to the expert knowledge related to the learning topics. |
| **Learnt Knowledge** | refers to the knowledge the agent acquired from the student tutor. |
| **Runtime Data** | refers to the data generated during runtime, such as, event information and history of events. |

After redefining the PTA, the improved PTA model will be introduced in section 3.2 ~ section 3.5.

## 3.2 Modelling the Main Routine of PTA

Section 3.2.1 will first introduce the methodology used to model PTA. Section 3.2.2 will then focuses on the modelling of the PTA's main routine using this methodology.

### 3.2.1 Goal Net Methodology

Goal Net methodology is well-suited for modelling goal-oriented agents. Since goal-orientation is one of the major characteristics of the PTA, Goal Net would be a suitable methodology to model the PTA.

Goal Net methodology is first proposed by Dr. Shen in 2005 [16] to model goal-oriented agents. In Goal Net, an agent pursues its goal by completing a sequence of sub-goals. A Goal Net can be composed from **four basic entities**: **states**, **transitions**, **arcs** and **branches**.

**States** are used to represent the goals of an agent. A state can be *atomic* or *composite*. A *composite state* can be further broken down into multiple *atomic states*. These *atomic states* represent the sub-goals that need to be fulfilled in order to fulfill the goal denoted by the *composite state*. States are represented by a round node in Goal Nets.

**Transitions** are the tasks that advance the agent from an *input state* to an *output state*. Each transition has an associate *task list*, which describes the *tasks* need to be completed in order to trigger this transition. Once all the *tasks* are completed, the transitions triggers and the agent advances from *input state* to *output state*. There are three types of transitions (as shown in Figure 3.2):

> **Direct Transition** The *task list* of such transitions is fixed and not changing according to runtime environment. The transition is triggered by completing same list of *tasks* every time.

**Conditional Transition** The *tasks* needed to be completed to trigger a conditional transition is selected dynamically during runtime.

**Probabilistic Transition** The *tasks* needed to be completed to trigger a probabilistic transition is selected dynamically based on probabilistic inference in an uncertain environment.

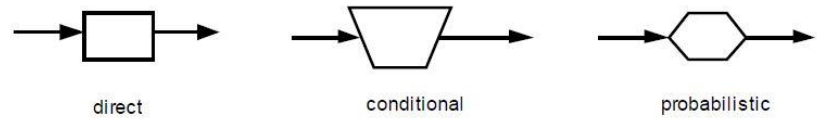

Figure 3.2 The Types of Transitions [16]

**Arcs** connect *states* and *transitions* and indicate the relationships between the nodes it connect. An arc can be represented by a *triangle arrow* or a *diamond arrow*.

**Branches** are used to denote *composite states*. Two branches connect a *composite state* to the sequence of sub-states that compose it. The left branch connects to the first sub-state of this composite state while right branch connected to the last sub-state.

Figure 3.3 shows an example of a Goal Net which contains all the four basic entities. S denotes *states*, while T denotes *transitions*.

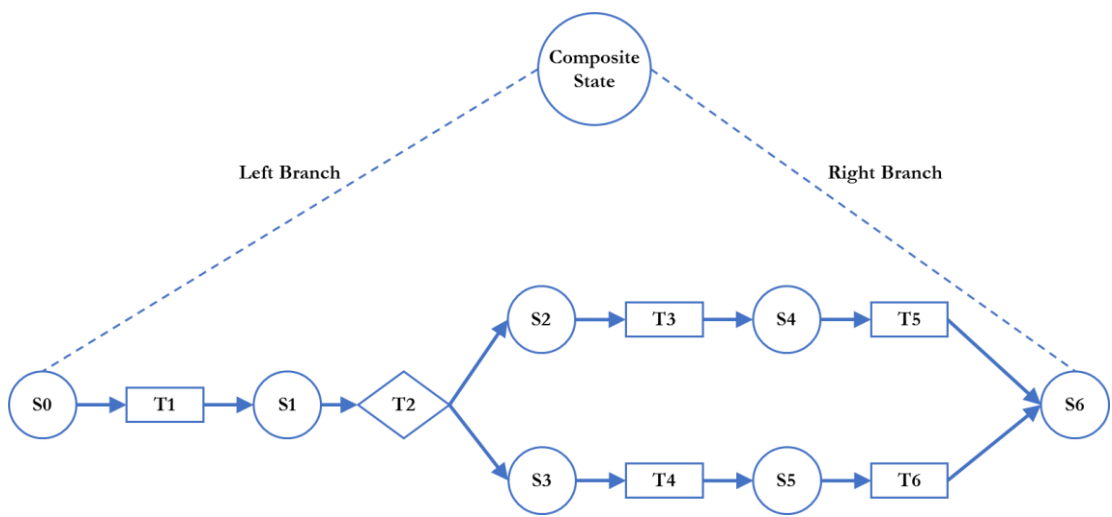

Figure 3.3 A Goal Net Example

## 3.2.2 Modelling PTA Main Routine with Goal Net

Main routine is the highest-level Goal Net model of the PTA (Figure 3.4), as it depicts the agent's pursuit of its root goal.

The design of main routing echoes the definition of the PTA in section 3.1. There are correspondences between the states and transitions of the main routine and the components of the PTA model, as shown in Table 3.1.

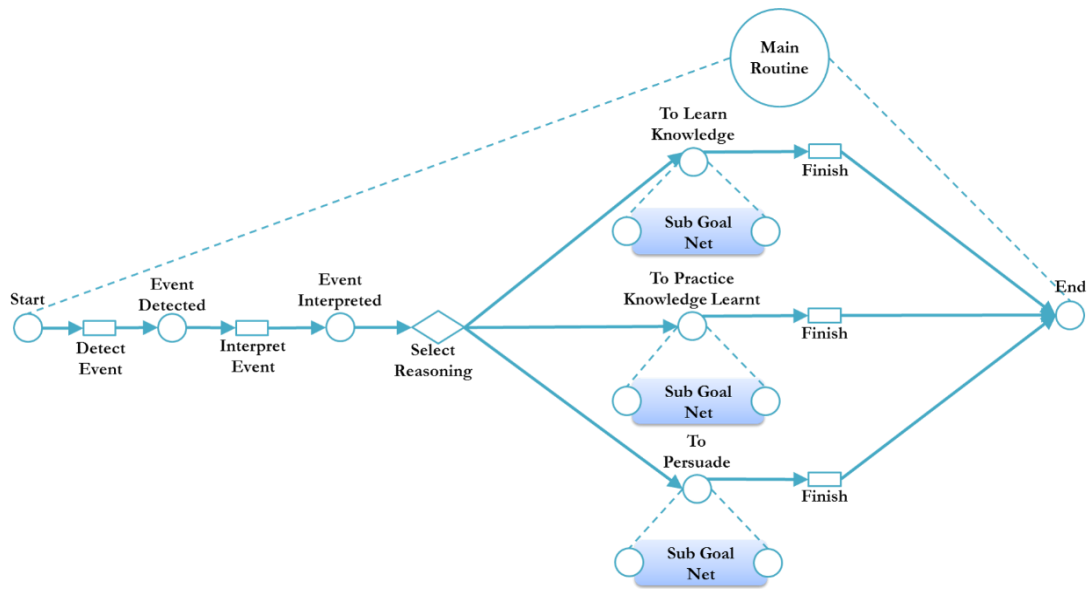

Figure 3.4 Goal Net Model of PTA

Table 3.1 Correspondences between the PTA Main Routine and the PTA Model

| States/Transitions in Main Routine | Component in PTA Definition |
| --- | --- |
| Detect Event, Event Detected, Interpret Event, Event Interpreted | **Et** Event Tracker |
| Selected Reasoning | **Rs** Reasoning Selection |
| To Learn Knowledge (and its sub Goal Net) | **Tr** Teachability Reasoning |
| To Practice Knowledge Learnt (and its sub Goal Net) | **Pr** Practicability Reasoning<br>**A** Actuator |
| To Persuade (and its sub Goal Net) | **P** Persuasion<br>**A** Actuator |

In the main routine, correspondences can also be found for the three cycles of the PTA defined in **Definition 3**. From the start node, executing the "To Learn Knowledge" branch to the end node would complete a teaching cycle. Executing the "To Practice Knowledge Learnt" branch to the end node would complete a practicing cycle. Executing the "To Persuade" branch to the end node would complete a persuasion cycle.

The main routine is executed repeatedly, so that the agent is constantly pursuing its goal. In each cycle, the agent checks for events repeatedly until one or more events are detected. According to the event(s) perceived, the agent selects an appropriate reasoning mechanism and executes the corresponding sub Goal Net of the selected reasoning. After the current cycle finishes, the agent reloads the start node of the main routine and begins with the next cycle.

The three sub Goal Nets will be described in details in the following sections.

## 3.3 Persuasion Reasoning of PTA

Section 3.3.1 will explain the persuasion theory which forms the basis of the Persuasion Reasoning. Section 3.3.2 will then explain the quantitative model used for the computational processes in the Persuasion Reasoning. Finally, section 3.3.3 will proposed the Goal Net model of the Persuasion Reasoning.

### 3.3.1 Persuasion Theory of PTA – ELM

The persuasion theory that forms the basis of the Persuasion Reasoning is the Elaboration Likelihood Model (ELM).

The Elaboration Likelihood Model (ELM) of persuasion is a dualistic model developed by Petty and Cacioppo in 1986. The ELM proposed an underlying framework for persuasion communication and the resulting attitude changes. Elaboration in the ELM refers to the process which the recipient of the persuasive message engages in cognitive thinking related to the persuasion topic. Various factors that can affect the likelihood of elaboration can be classified into two large categories: motivation of the message recipient and his or her ability to elaborate.

According to the ELM, there are two routes of persuasion, the central route and the peripheral route [14], as shown in Figure 3.5. The likelihood of elaboration determines which route the message recipient would go through. The likelihood of elaboration increases if one is motivated and has the ability to engage in elaboration. In this situation, one is likely to go through the central route of persuasion where his or her attitude changes tend to be enduring. On the contrary, the likelihood of elaboration decreases if one is not motivated or does not have ability to elaborate. In this situation, one is likely to go through the peripheral route of persuasion where the resulting attitude changes tend to be less enduring.

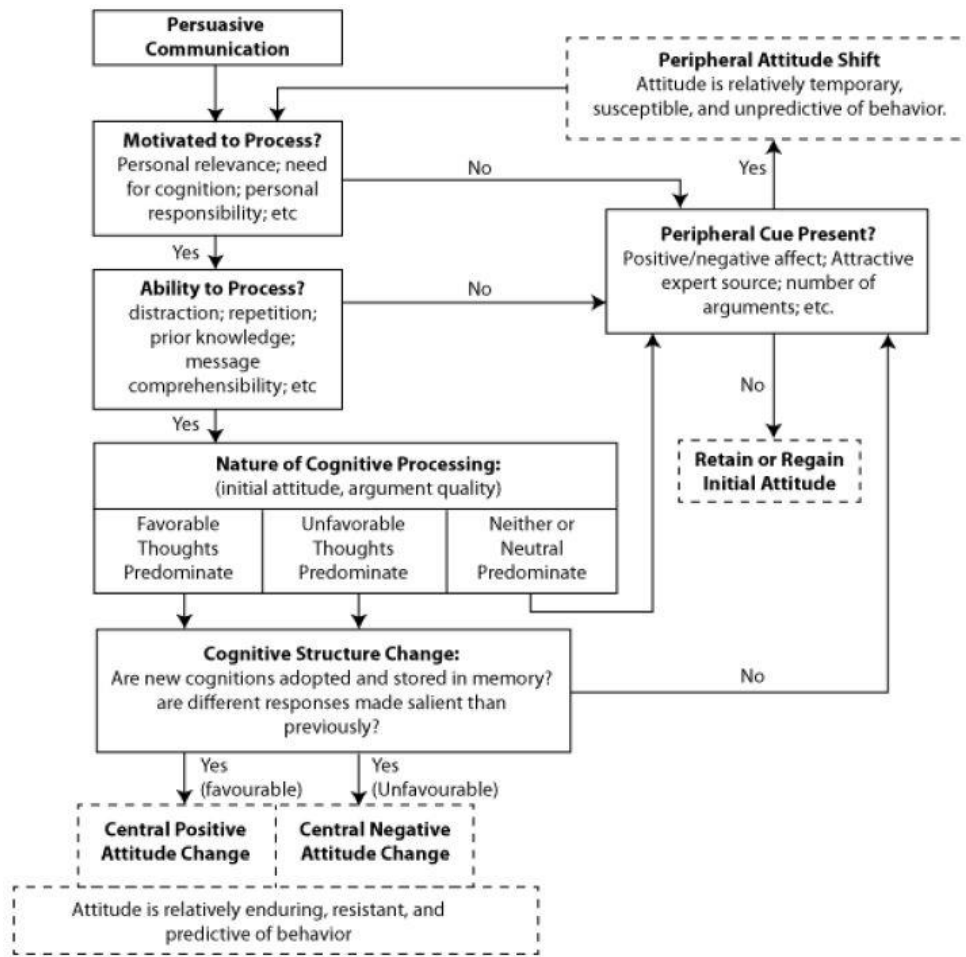

Figure 3.5 Central and Peripheral Route of Persuasion

The following are the factors [14] that can affect the likelihood of elaboration explained in the context of PTA, arranged according to the categories they belong to.

**Category 1: Motivation.** Motivation can be assessed from following factors:

(1) *Personal Relevance.* How relevant the event is to the learning topics.

(2) *Personal Responsibility.* Whether the event demonstrates the student's responsibility towards his or her own learning.

(3) *Need for Cognition.* Whether the student enjoys the learning process. When the learning process is enjoyable, the student would be more willing to engage in cognitive thinking.

**Category 2: Ability.** Ability can be assessed from following factors:

(1) *Prior Knowledge.* Whether the student has acquired the necessary knowledge required to teach the PTA.

(2) *Distraction.* Whether the event is a distractor and would decrease the student's ability.

(3) *Repetition.* Whether the student is revisiting the knowledge he or she has acquired by teaching the PTA.

The PTA has incorporated the ELM to achieve its goal of instilling enduring positive attitude changes and knowledge gains of the student, i.e. keep him or her on the central route of persuasion. By assessing the six factors mentioned above, the PTA can evaluate the motivation and ability level of the student and generate appropriate persuasion cue if the student is in low motivation or low ability state.

**3.3.2 Quantitative Model for Assessing Motivation and Ability – FCM**

As mentioned above, the PTA needs to determine the motivation and ability level of the student in order to generate appropriate response. FCM can be applied to build a quantitative model to derive the values of motivation and ability from the values of their respective constituent factors.

Fuzzy Cognitive Map (FCM) [15] is a fuzzy tool which can be used to depict causal relationships in dynamic systems. FCM is a graphical representation, at the same time, also a mathematical model of the dynamic system it is modelling.

A FCM is composed of *causal concepts* and *causal relationships*. A *causal concept* can be used to model an event, a goal or an action. It is represented by a rounded node in FCM and has a real number value in interval [-1, 1]. A set of *causal concepts* are interconnected by directed edges. Each directed edge represents the *causal relationship* between two *causal concepts*. A *causal relationship* with a positive sign "+" denotes casual increase, while one with a negative sign "-" denotes causal decrease. A *causal relationship* can have a weight associated with it to denote the strength of the cause-effect relationship it represents. The weight usually takes a real value in interval [-1, 1].

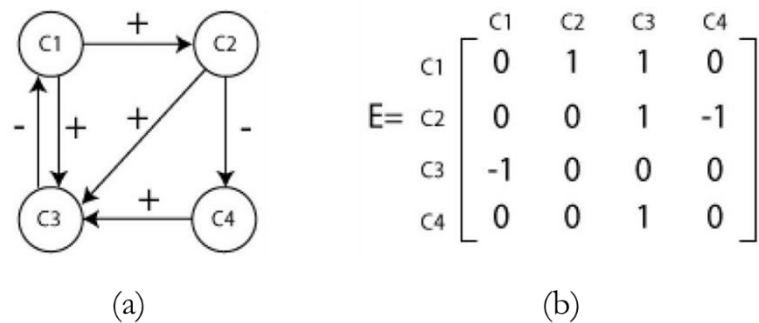

(a)          (b)

Figure 3.6 A Simple FCM Example

Figure 3.6 shows a simple FCM example. The FCM in Figure 3.6(a) can also be represented as a matrix (denoted by E) in Figure 3.6(b). The weight of causal relationship from $C_i$ to $C_j$ is the value at $E_{ij}$. For example, the weight of causal relationship from $C_2$ to $C_4$ is -1, which means $C_2$ decreases $C_4$.

In the context of a PTA, FCM can be applied to build a quantitative model to derive the values of motivation and ability from the values of their respective constituent factors. A FCM model proposed for the PTA is shown in Figure 3.7. This FCM model depicts the causal relationship among motivation, ability and factors affect them.

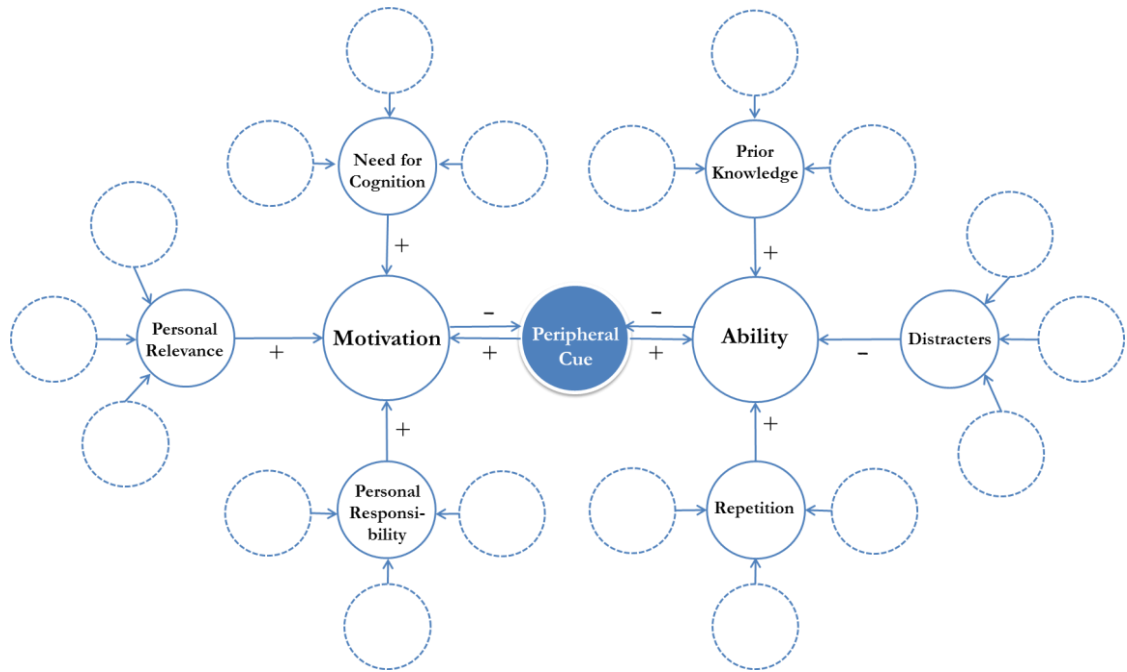

Figure 3.7 FCM Model of PTA

**Definition 5**: A **stem node** is a causal concept that is:

(1) Peripheral Cue or;
(2) Motivation or;
(3) Ability or;
(4) A factor of motivation or ability.

All PTA FCMs have the same *stem nodes*.

**Definition 6**: A **leaf node** is a causal concept that is not a *stem node* and has a causal relationship with one of the factors of motivation or ability.

*Leaf nodes* model the events that have causal relationships with the motivation or ability factors. Thus, the type and number of *leaf nodes* differ from one PTA implementation to another.

Multiple *leaf nodes* can be connected to a motivation or ability factor. The same *leaf node* can also be connected to multiple factors. A *leaf node* is denoted by a rounded node with dotted rim in Figure 3.7.

**Definition 7**: the **Main FCM** is the FCM formed by all the *stem nodes*.

In Figure 3.7, all the nodes with solid rim form the **Main FCM**.

**Definition 8**: a **Sub FCM** is a FCM formed by a *stem node* and all the *leaf nodes* that are connected to it.

In the PTA FCM, there are six *Sub FCMs*, each formed around a factor of motivation or ability.

With the proposed PTA FCM model and the above definitions, the computational processes in the Persuasion Reasoning can be designed with greater flexibility and implemented with greater computational efficiency. An implementation of the proposed PTA FCM model is described in section 5.2.4.

### 3.3.3 Modelling Persuasion Reasoning with Goal Net

The sub Goal Net of the Persuasion Reasoning (Figure 3.8) depicts how the PTA pursues its sub goal, "To Persuade". The ability to persuade is one of the important features that characterize a PTA. In pursuit of its sub goal "To Persuade", a PTA would act to keep the student on the central route of persuasion, i.e. to keep the student in high motivation and high ability state.

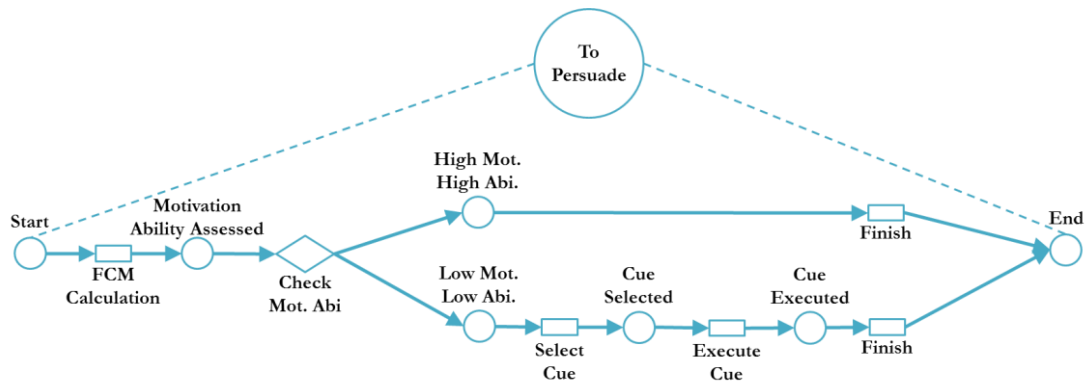

Figure 3.8 Goal Net Model of Persuasion Reasoning

When this sub Goal Net is executed, firstly, the agent assesses the values of motivation and ability according to the Fuzzy Cognitive Map Model proposed in section 3.3.2. Secondly, the agent determines whether the resultant motivation and ability values are high or low by comparing them to the pre-determined baseline values. If both motivation and ability values are high, no particular action is required. However, when one of the values is low, the agent selects an appropriate persuasion cue and executes it.

After execution of the sub Goal Net finishes, the agent continues to execute the next node after the composite state "To Persuade" on the main routine.

## 3.4 Teachability Reasoning of PTA

### 3.4.1 Integration with Practicability and Persuasion

The integration point between the Teachability Reasoning and the Practicability Reasoning is the knowledge base. In a teaching cycle, the PTA learns from the student and saves the knowledge learnt into the knowledge base. In subsequent practicing cycle, the Practicability Reasoning reads the saved knowledge.

The Teachability Reasoning also needs to be integrated with the Persuasion Reasoning. During the teaching cycle, any indication that suggests the student is low in motivation or ability will be captured and signaled, so that the Persuasion Reasoning can process it in a persuasion cycle later.

### 3.4.2 Modelling Teachability Reasoning with Goal Net

The sub Goal Net of the Teachability Reasoning (Figure 3.9) depicts how the PTA pursues its sub goal, "To Learn Knowledge".

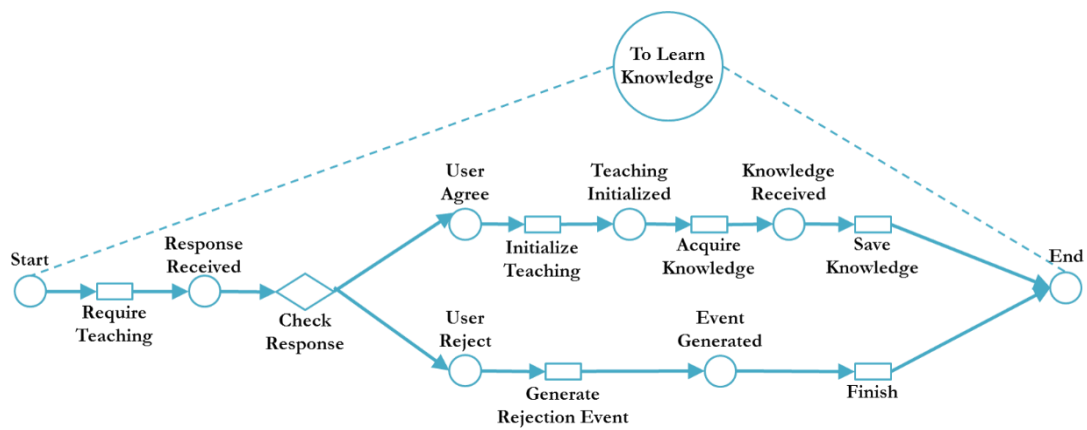

Figure 3.9 Goal Net Model of Teachability Reasoning

When executing this sub Goal Net, firstly, the agent requires the student to teach it. Secondly, it waits for and checks the response from the student. If the student agrees to teach the PTA, the PTA will learn from the student and save acquired knowledge to the knowledge base.

However, if the student is not motivated to or does not have the ability to teach the PTA, he or she would refuse the teaching request from the PTA. In this case, the PTA will generate a rejection event. This event will be detected by the PTA in the next cycle. As the refusal indicates that the student has either low motivation or low ability, the agent will choose to go through persuasion reasoning during when a rejection event is detected.

After execution of this sub Goal Net finishes, the agent continues to execute the next node after the composite state "To Learn Knowledge" on the main routine.

## 3.5 Practicability Reasoning of PTA

### 3.5.1 Integration with Teachability and Persuasion

The Practicability Reasoning needs to be integrated with the Teachability Reasoning in order to generate teaching feedback. In a case when the student tutor commits errors during teaching, the errors should be captured and signaled during Practicability Reasoning, so that the Teachability Reasoning can highlight the errors during next teaching cycle.

The Practicability Reasoning also needs to be integrated with the Persuasion Reasoning. During the practicing cycle, any event that may lead the student to a low motivation or low ability state will be captured and signaled, so that the Persuasion Reasoning can process it in a persuasion cycle later.

### 3.4.2 Modelling Practicability Reasoning with Goal Net

The sub Goal Net of the Practicability Reasoning (Figure 3.10) depicts how the PTA pursues its sub goal, "To Practice Knowledge Learnt".

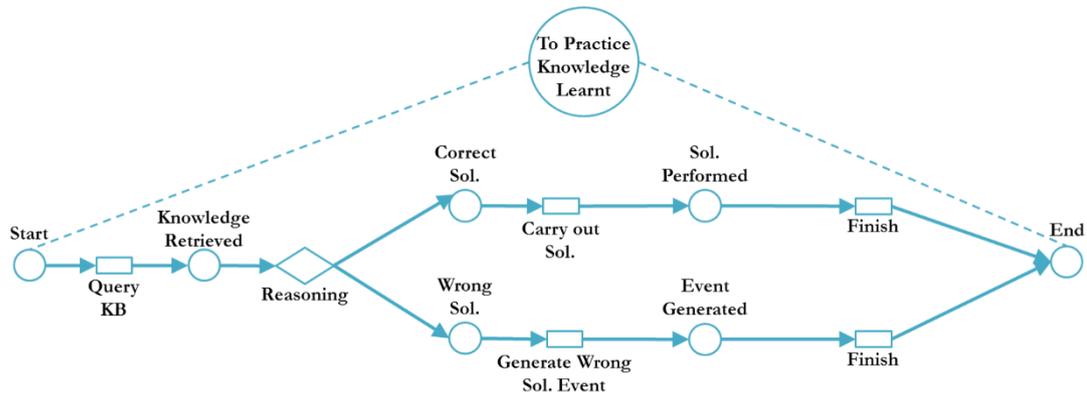

Figure 3.10 Goal Net Model of Practicability Reasoning

When executing this sub Goal Net, the agent first queries the knowledge base to retrieve the knowledge it has acquired from the student. Secondly, it derives a solution from the knowledge by reasoning. If the resultant solution is correct, it will carry out the solution.

However, if the resultant solution is wrong, the agent will generate a wrong solution event, which will be detected by the agent in the next cycle. As the student may be demotivated after this fail trail, the PTA will choose to go through the Persuasion Reasoning when a wrong solution event is detected.

After execution of this sub Goal Net finishes, the agent continues to execute the next node after the composite state "To Practice Knowledge Learnt" on the main routine.

# Chapter 4 System Architecture Design

## 4.1 Overall System Architecture Design

In Chapter 3, an improved agent model for the PTA is proposed. However, since the agent model has a high level of abstraction, there is still a gap between this theoretical model and the implementation of the PTA. In order to provide more guidelines for implementation, new system architecture for the PTA is proposed in this chapter.

The proposed system architecture (as shown in Figure 4.1) focuses on the design of the agent's **control structure**. The system architecture retains certain level of abstraction and is reusable as:

(1) It is independent of the learning topics, the deployment environment and the form of embodiment of the PTA;

(2) It guides, but does not restrict the way of implementing a PTA.

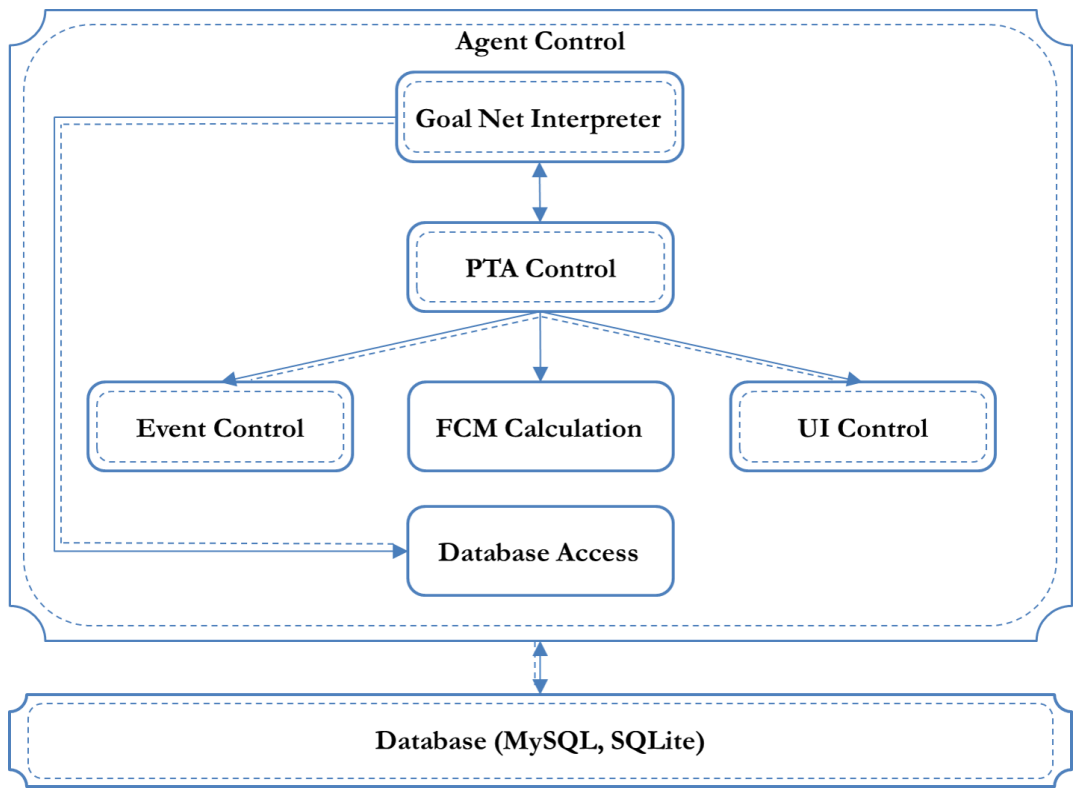

Figure 4.1 System Architecture for PTA

At the highest level, the Goal Net Interpreter directs the goal-oriented behaviors of a PTA according to its Goal Net models. It determines the next state of the PTA and the tasks to be performed in order to advance to the next state. The Goal Net Interpreter instructs the PTA Control component to carry out the tasks required. The PTA control is supported by a number of other control components. It dispatches sub-tasks to Event Control, FCM Calculation and UI Controls and coordinates the execution of the sub-tasks. Table 4.1 provides a summary of the roles and functionalities of each component.

Table 4.1 Functionalities/ Roles of Components in System Architecture

| Component | Functionality/ Role |
| --- | --- |
| Goal Net Interpreter | Directs the goal-oriented behaviors of a PTA |
| PTA Control | Acts as a central dispatcher and coordinator |
| Event Control | Deals with the creation, logging, processing and removal of events |
| FCM Calculation | Performs computational processes related to FCM |
| UI Control | Control the interface elements in the PTA environment |
| Database Access | Connects to the database and performs queries |

In the following sections, each component of the system architecture will be explained in greater details in terms of:

(1) Its main functionalities;

(2) Its interactions with other components;

(3) Possible ways to implement this component.

## 4.2 Goal Net Interpreter

The Goal Net Interpreter directs the goal-oriented behaviors of a PTA according to the Goal Net Models introduced in Chapter 3. As shown in Figure 4.2, the proposed Goal Net models of the PTA can be created by drawing using a graphical Goal Net Designer [18] and saved into the Goal Net database. Please refer to **Appendix 2 for** the complete Goal Net model drawn in a Goal Net Designer.

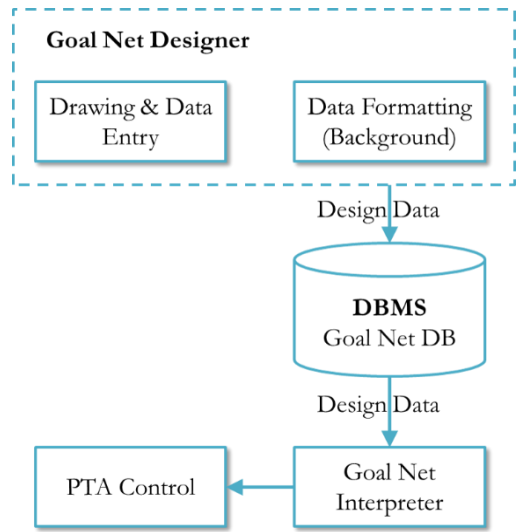

Figure 4.2 Designing and Using the Goal Net Models

The Goal Net Interpreter loads the Goal Net models from the database during runtime and interprets them dynamically. It determines the next state of the PTA by traversing through the nodes in the PTA's Goal Net. During a state transition, the Goal Net Interpreter also determines the tasks the PTA should perform in order to pursue its goal.

After determining the tasks to be performed, the Goal Net Interpreter instructs the PTA Control to perform those tasks. After finishing the designated tasks, the PTA Control informs the Goal Net Interpreter to proceed to the next state.

There are two possible ways to implement the Goal Net Interpreter: to make use of an existing Goal Net interpreter or to build a new one.

**Alternative 1**: *Make use of an existing Goal Net interpreter.* MADE Runtime [18] is a virtual machine that interprets the Goal Net models. It loads the Goal Net designs from the database using a Goal Net loader (as shown in Figure 4.3) and traverses through the Goal Net. Whenever a task needs to be performed to fire a state transition, MADE Runtime will invoke the corresponding routine. In this case, MADE Runtime is running as an active component and the system architecture would be different from what is described in Figure 4.1. The system architecture using MADE Runtime is shown in Figure 4.4.

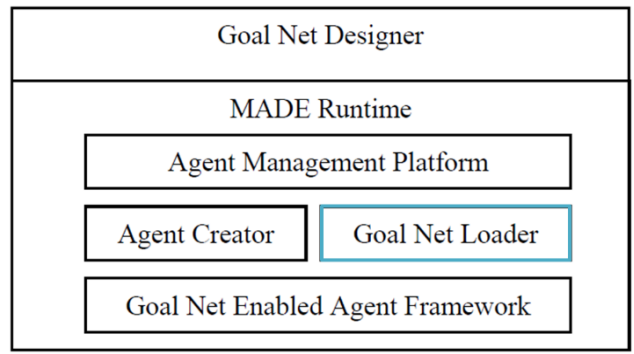

Figure 4.3 Architecture of MADE Runtime [18]

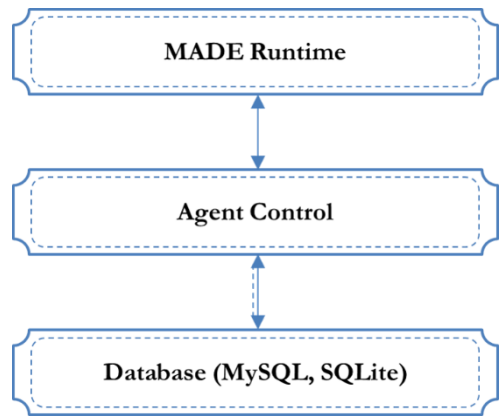

Figure 4.4 System Architecture with MADE Runtime

**Alternative 2**: *Build a new Goal Net Interpreter.* The most direct way to build a new Goal Net Interpreter is to directly access the Goal Net database. Goal Net database stores the information of states, transitions, task lists and functions. A Goal Net Interpreter can be implemented by querying the above information directly to determine the next state of the PTA and the tasks to be performed. Figure 4.5 shows the logic of how the Goal Net Interpreter pursues the goal of the agent. In order to access the Goal Net Database, the Goal Net Interpreter needs to use the interfaces provided by the Database Access component.

```
Load initial state
WHILE current node is not goal state
    IF current node is a transition node
        IF current node is a decision node
            Find all the states following this decision node
            Choose a state from the states found as the next state
        ELSE IF current node is a normal transition node
            Find the next state following it

        IF next state is goal state
            Set the WHILE condition to false

        IF current node has a task list
            Get all the functions in the task list
            FOR each of the functions
                Invoke the function

        Update current state to next state

    ELSE
        Find and go to the next node
```

Figure 4.5 Pseudo Code for Work Flow of Goal Net Interpreter

## 4.3 PTA Control

The PTA Control component implements the major functionalities of a PTA. While the Goal Net Interpreter directs the agent's goal-oriented behaviors, the PTA Control coordinates them.

The Goal Net Interpreter instructs the PTA Control to perform certain tasks by invoking corresponding task functions in the PTA Control. While executing the task functions, the PTA Control breaks down the tasks into sub-tasks and dispatches them to others components by calling respective modules in those components. In this manner, all the components work towards the goal of the PTA collectively.

In order to realize all the three cycles of a PTA defined in **Definition 3**, the PTA Control needs to implement the task functions listed in Table 4.2. Each of these task functions is associated with a transition in the Goal Net models of the PTA. A task function has the same name as the transition it is associated to.

Table 4.2 Task Functions to be Implemented in the PTA Control

| Goal Net Model | Functions to be Implemented | |
| --- | --- | --- |
| **Main Routine** | DetectEvent | SelectReasoning |
|  | InterpretEvent | Finish |
| **To Learn Knowledge** | RequireTeaching | SaveKnowledge |
|  | CheckResponse | GenerateRejectionEvent |
|  | InitializeTeaching | Finish |
|  | AcquireKnowledge |  |
| **To Practice Knowledge Learnt** | QueryKB | GenerateWrongSolEvent |
|  | Reasoning | Finish |
|  | CarryOutSol |  |
| **To Persuade** | FCMCalculation | ExecuteCue |
|  | CheckMotAbi | Finish |
|  | SelectCue |  |

## 4.4 Other Components Supporting PTA Control

The PTA Control coordinates the goal-oriented behaviors of the PTA and implements the major functionalities of the PTA. Event Control, FCM Calculation and UI Control are three components that support the PTA Control. The three of them deals with the percepts, reasoning and actions of the agent respectively. The Event Control component enables the agent to perceive its environment. The FCM Calculation enables the agent to reason about the motivation and ability level of the student. The UI Control component enables the agent to act on its environment.

**4.4.1 Event Control**

The Event Control component corresponds to the Event Tracker **Et** (refer to definition in section 3.1) of the PTA. Event Control handles the lifetime of an event. It deals with the creation, logging, processing and removal of events.

The PTA Control makes use of the functions in Event Control to handle events.

It is more recommended to implement the Event Control using the event checking mechanism with a checking period of a few seconds. During the checking period, the newly generated events are logged in the Event Log of the Event Control. When checking is performed, the Event Control prioritizes the events and decides the processing sequence of the events. It may also decide to process a number of interactional events in a batch.

Table 4.3 suggests a few types of events that can be tracked in a 2D or 3D PTA environment, together with corresponding examples.

Table 4.3 Suggested Event Types to be Tracked in a PTA Environment

| Event Category | Event | Example |
| --- | --- | --- |
| **Learning Behavior** | Dialogue Event | a. The student has refused to teach the PTA in a dialogue with the PTA<br>b. The student talks to a distracting character, such as a rabbit |
| | Location Event | The student has arrived at a designated location |
| | Time Event | The student has been inactive for more than 5 minutes |
| **Learning Achievement** | Collection of Item | The student has collected a designated item |
| | Fulfillment of Mission | The student has completed a learning task |
| **Knowledge Data** | Commitment of Errors | The student commits an error while teaching the PTA |
| | Teaching Feedback Event | The PTA cannot generate correct solution from the knowledge acquired from the student – wrong solution event |

### 4.4.2 FCM Calculation

The FCM Calculation component performs the computational processes related to FCM in the Persuasion Reasoning of the PTA.

When the PTA is initialized, the FCM model is loaded from the database and stored in the FCM Calculation component. In the Persuasion Reasoning cycle, the PTA Control invokes the functions in the FCM Calculation to calculate the motivation and ability value of the student.

There are two possible ways to implement the FCM Calculation component: to leverage on existing FCM modelling tools and software or to build a new one from scratch.

**Alternative 1**: *Leverage on existing FCM modelling tools and software*. There are quite a number of FCM tools available for free. An interface can be built for these tools so they can be used by the PTA control structure. The interface needs to handle the conversion and passing of the inputs to these tools and the retrieval and reformatting of the outputs from them. This alternative would greatly reduce the effort required to build the FCM Calculation component. However, most of the FCM tools available are not easy to integrate. Table 4.4 summarizes a few popular FCM tools and some of their disadvantages.

Table 4.4 Summary of Popular FCM tools

| Tool Name | Descriptions | Disadvantages |
|---|---|---|
| **FCMappers** | ▪ Based in Excel<br>▪ Visual representations for FCM on the web | ▪ Not easy to integrate<br>▪ Cannot customize parameters or threshold functions of FCM |
| **Mental Modeller** | ▪ Downloaded or web-based<br>▪ GUIs for drawing FCMs<br>▪ Interface for display FCM matrix | ▪ Not easy to integrate<br>▪ Cannot customize parameters or threshold functions of FCM |
| | | ▪ Limited causal relationship values (can only choose from a few values) |
| **JFCM** | ▪ Downloaded Java Program<br>▪ Creating FCMs by writing a piece of Java code or reading from XML files<br>▪ Open source | ▪ Not customized for PTA FCM, thus does not reflect the structure of PTA FCM |

**Alternative 2**: *Build a new FCM tool from scratch*. Building a new one would require more effort. However, this alternative allows more space for customization and greatly reduces the difficulty of integrating with other components. An implementation of the proposed PTA FCM model is described in section 5.2.4.

**4.4.3 UI Control**

The UI Control component corresponds to the Actuator **A** (refer to definition in section 3.1) of the PTA. It contains modules that enable the agent to act on various aspects of its environment, including the visual representation of itself in the environment.

For example, the PTA can be embodied into its environment as 2D figure or as a 3D avatar. The UI Control provides modules that can control the figure or avatar of the PTA. The UI Control can also control the graphical interfaces or visual effects in the PTA environment. Further examples can be found in section 5.4.3.

## 4.5 Database Access

The Database Access component provides interfaces for the other components to access the database.

Depends on the implementation of the PTA, multiple databases may need to be accessed. For example, the Goal Net Interpreter needs to access the state and transition information in Goal Net database, while the FCM Calculation component needs to load the FCM model from the FCM database.

The Database Access component should be able to support and coordinate access to multiple databases. It should provide modules to connect to different database management systems where the databases are managed. When two different databases are managed by the same database management system, the Database Access component needs to coordinate the access to these two databases as the best practice is to keep only one open connection at any time.

# Chapter 5 Implementation of PTA

## 5.1 Game Environment

### 5.1.1 Unity for 3D Game Development

Unity is a development platform for creating 2D/3D games and interactive experiences. Unity includes a game engine and an integrated development environment.

VS Saga is a 3D video game developed for this project on the Unity development platform. An instance of the PTA is embedded in VS Saga to demonstrate the practice of instantiate a PTA from the propose agent model and system architecture. Please refer to **Appendix 5** for the complete class diagram for the agent control structure.

### 5.1.2 Main Scenes and Story Lines

VS Saga is a pedagogical game developed to facilitate the learning of secondary school science. Following the PTA model and system architecture proposed, a PTA, called "Water Molecule", is developed and embedded in the game.

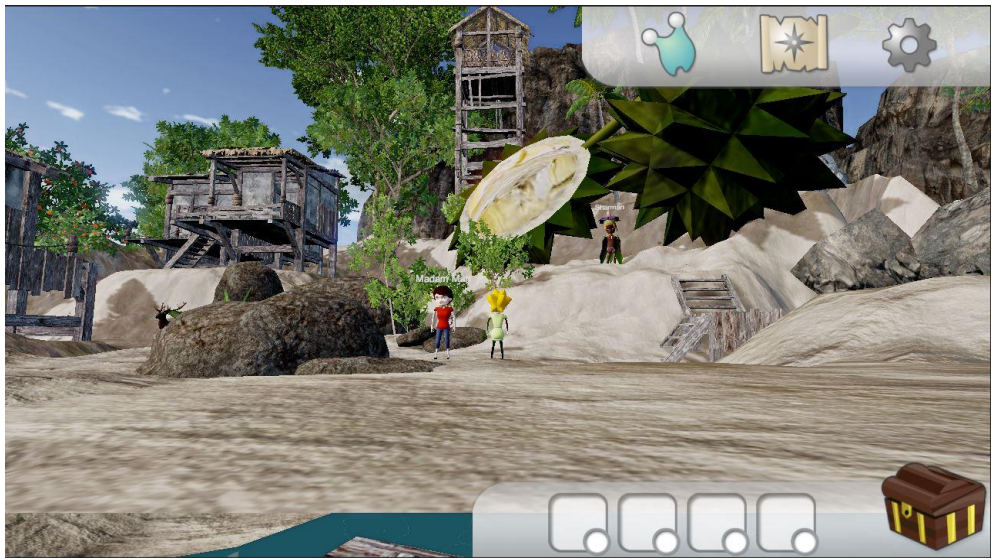

Figure 5.1 Interface of VS Saga

In VS Saga, the student learns how water molecules are transported within a plant by completing a mission. The mission of the student is to help the water molecule enter the root of a banana tree. The student learns about the concept of diffusion and osmosis and then **teaches** the PTA, the water molecule, these concepts to help it enter the root.

There are three scenes in VS Saga, the Knowledge Town, the Science Laboratory and the Tree. In the following part of this section, the story line and main characters of each scene are introduced.

**Scene 1: the Knowledge Town**. The knowledge town is a small island where the student meets the villagers and learns about the concept of diffusion and osmosis from them.

| Screen Shot | Story Line and Main Characters |
|---|---|
| 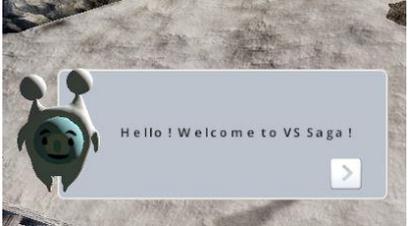<br>Figure 5.2(a) Water Molecule | 1. Meet the Water Molecule (**the PTA**)<br>    1.1 Welcomed by the Water Molecule<br>    1.2 Water Molecule introduces itself<br>    1.3 Water Molecule explains the mission of this game |
| 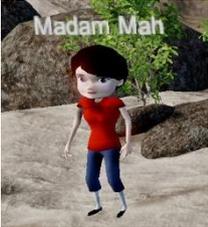<br>Figure 5.2(b) Madam Mah | 2. Meet Madam Mah on the beach<br>    2.1 Welcomed by Madam Mah<br>    2.2 Accept a new quest "**Meet Sharman**" |
| 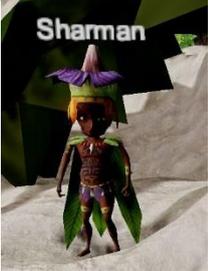<br>Figure 5.2(c) Sharman | 3. Meet Sharman near the durian tree<br>    3.1 **Learn diffusion** from Sharman<br>        3.1.1 Accept a new quest "**Get Perfume from Madam Sammy**"<br>    3.2 Refuse to learn diffusion |

| Screen Shot | Story Line and Main Characters |
|---|---|
| 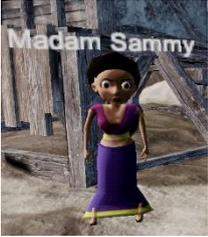<br>Figure 5.2(d) Madam Sammy | 4. Meet Madam Sammy near the beach house<br>    4.1 Reminded by Madam Sammy to learn diffusion if have not done so<br>    4.2 Get a new item "Potion"<br>    4.3 Get a new quest "**Pass Potion to Mayor**" |
| 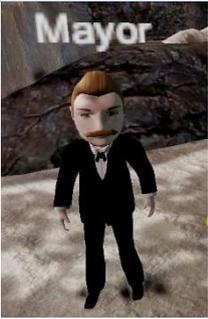<br>Figure 5.2(e) Mayor | 5. Meet the Mayor under the coconut tree<br>    5.1 **Learn osmosis** from the Mayor<br>        5.1.1 Invited by the Mayor to the Science Laboratory<br>    5.2 Refuse to learn osmosis |

**Scene 2: the Science Laboratory**. The Science Laboratory is another small island where the student practices the knowledge he or she has learnt. There is a big diffusion tank in the science laboratory where the student can conduct diffusion and osmosis experiments.

| Screen Shot | Story Line and Main Characters |
|---|---|
| 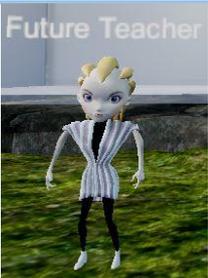<br>Figure 5.3(a) Future Teacher | 1. Meet Future Teacher<br>    1.1 Learn how to conduct experiments from the Future Teacher<br>        1.1.1 Agree to conduct experiments<br>        1.1.2 Refuse to conduct experiments |

| Screen Shot | Story Line |
|---|---|
| 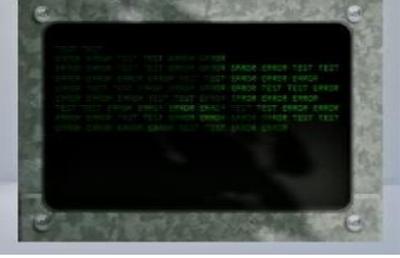<br>Figure 5.3(b) Diffusion Tank Panel | 2. Proceed to the Control Panel of the Diffusion Tank, click to start the experiments |
| 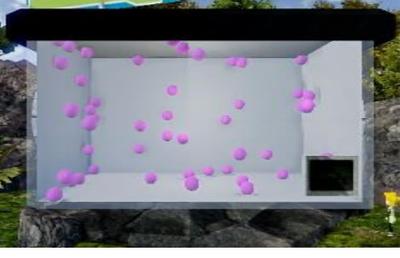<br>Figure 5.3(c) Diffusion Experiment | 3. Conduct the diffusion experiment |
| 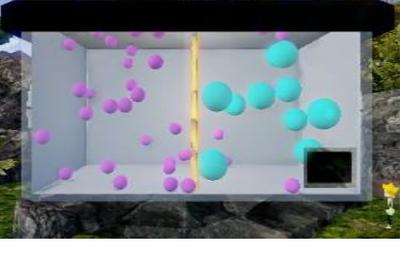<br>Figure 5.3(d) Osmosis Experiment | 4. Conduct the osmosis experiment<br>   4.1 After finishing the two experiments, told by the Future Teacher to proceed to Scene 3 to save the water molecule |

**Scene 3: the Tree**. The banana tree is on the third island. This is where the student teaches the PTA (the water molecule) about diffusion and osmosis. The PTA practices the knowledge learnt by trying to enter the root of the tree.

| Screen Shot | Story Line and Main Characters |
|---|---|
| 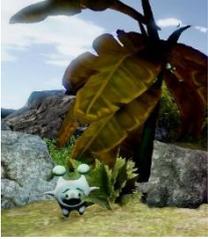<br>Figure 5.4(a) Withered Tree | 1. Arrive at the banana tree, see the withered tree and sad water molecule (**the PTA**) |
| 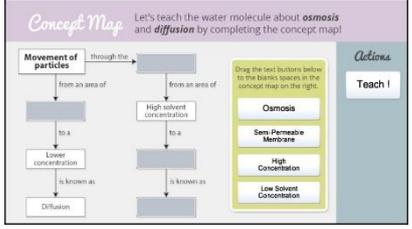<br>Figure 5.4(b) Concept Map | 2. Receive teaching request from the PTA<br>　　2.1 Agree to teach the PTA by completing<br>　　　　a concept map (Figure 5.4(b))<br>　　2.2 Refuse to teach the PTA |
| 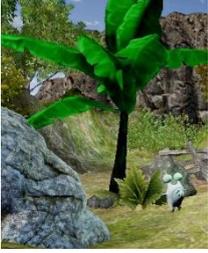<br>Figure 5.4(c) Revitalized Tree | 3. The PTA practices the knowledge learnt<br>　　3.1 Water molecule successfully enters the<br>　　　　root and the tree is revitalized (Figure<br>　　　　5.4(c))<br>　　3.2 Water molecule fails to enter the root<br>　　　　and requires teaching again |

Except for the main characters, there are other distracting characters in the game environment, such as a rabbit and a stag.

## 5.2 Goal Net Interpreter

Since there is no existing Goal Net interpreter for Unity, a new one is implemented according to the logic described in Figure 4.5. The implemented Goal Net Interpreter directly accesses the Goal Net database using the interface provided by the Database Access Component.

The Goal Net Interpreter advances from an input state to an output state by firing the transition that connects them. However, when it encounters a decision node for which one input state has multiple possible output states, it needs to decide which output state to move to. To resolve that issue, a decision table is implemented in the PTA Control component (as shown in Figure 5.5) which keeps the decision to be made at each decision node. The PTA Control updates this decision table according reasoning results and current percepts. The Goal Net Interpreter refers to the decision table for deciding the output state if a decision node is encountered.

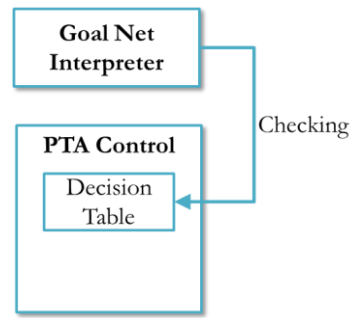

Figure 5.5 Decision Making in Goal Net Interpreter

For each transition, the Goal Net Interpreter retrieves all the tasks in the task list of the transition and invokes the corresponding task functions in the PTA Control using their function names. The code snip for calling a task function using its function name is shown in Figure 5.6.

```
// Performing actions calls
try {
    Type T = typeof(ManagerScript_GameController);
    T.GetMethod (currentAction).Invoke (game, null);
} catch (NullReferenceException e) {
    Debug.Log ("Cannot get type from GameController instance.");
} catch (ArgumentNullException e) {
    Debug.Log ("Argument for invoke function is null.");
}
```

Figure 5.6 Invoking a Function with Function Name

## 5.3 PTA Control

The PTA Control has implemented the task functions in Table 4.2. This section provides summaries of how these functions are implemented in the context of VS Saga.

**Main Routine**. During the execution of the PTA's Main Routine, the PTA Control calls the Event Control component every 5 seconds to check for events until one or more events are detected. Then, it prioritizes the events and decides the ones to be processed in the current cycle. It determines which the reasoning cycle the event(s) should go through and updates the decision table.

**Persuasion Reasoning**. During the execution of the sub Goal Net "To Persuade", the PTA Control calls the FCM Calculation component to calculate the value of motivation and ability based on current percepts (events). Then, it compares the resultant values to the baselines values to determine whether the student is in low motivation or low ability state where persuasion is needed. It updates the decision table accordingly. When persuasion is needed, the PTA Control selects a suitable cue according to the attributes of the event(s). Lastly, the PTA Control calls the UI Control to display the selected persuasion cue (an example in Figure 5.7).

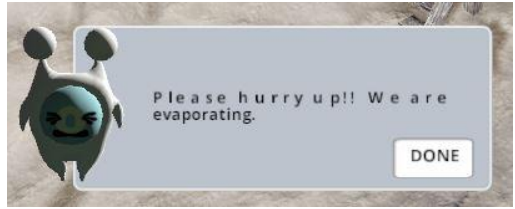

Figure 5.7 Example of a Persuasion Cue

**Teachability Reasoning**. As shown in Figure 5.4(b), the teaching process is mainly performed on a UI called Concept Map [19], which shows the relationships between the key concepts of the learning topic. The water molecule requests teaching from the student during a dialogue, as shown in Figure 5.8. If the student agrees to teach the water molecule, the PTA Control will display the concept map. After teaching process finishes, the knowledge taught by the student will be stored within the script of the concept map. If the student refuses to teach the water molecule, the PTA control will generate a rejection event.

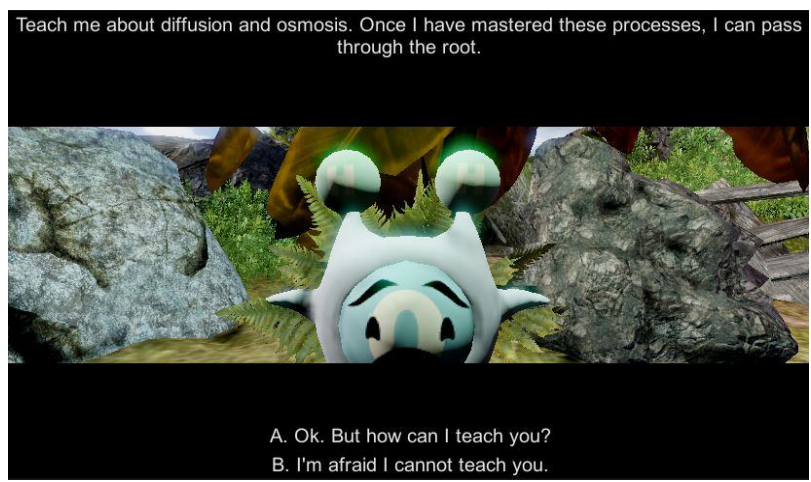

Figure 5.8 Water Molecule Requests Teaching from the Student

**Practicability Reasoning**. During the execution of the sub Goal Net "To Practice Knowledge Learnt", the PTA Control reasons the correctness of the teaching based on the knowledge stored in the concept map. If the teaching of the student is correct, the PTA Control will invoke the animation of the banana tree so the tree becomes revitalized as shown in 5.13. Otherwise, the PTA Control will generate a wrong solution event.

## 5.4 Other Controllers Supporting PTA Control

### 5.4.1 Event Control

In VS Saga, the Event Control component mainly tracks and deals with three types of events in Table 4.3: dialogue events, time events and teaching feedback events.

**Dialogue Events**. A dialogue event is an event created when a particular sentence in a dialogue has been spoken. A dialogue event can be created within the dialogue system of Unity using a special type of script called SequencerCommand.

The student player's dialogues with the game characters can be configured in the dialogue system. Figure 5.9 shows the dialogue entry of a sentence in the dialogue between the student and the water molecule in Unity. This sentence, "B. I'm afraid I cannot teach you", is spoken by the student to the water molecule to refuse the teaching request from it.

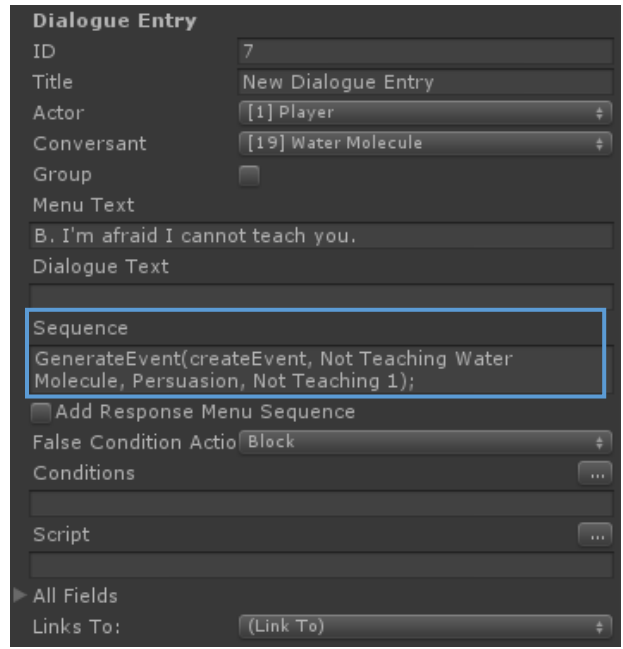

Figure 5.9 Example Dialogue Entry in Unity Dialogue System

The blue rectangle in Figure 5.10 highlights the SequencerCommand *GenerateEvent* which will be called when this sentence is spoken. Figure 5.10 shows the code snip of the SequencerCommand *GenerateEvent*. This SequencerCommand extracts the parameters passed to it and calls the function for creating a new event with the parameters.

```
public class SequencerCommandGenerateEvent : SequencerCommand {

    private ManagerScript_EventController eventController;

    void Start () {
        //eventController = ManagerScript_EventController.Instance;
        eventController = GameObject.Find ("Managers_SCN01").
                    GetComponent<ManagerScript_EventController> ();

        string functionName = GetParameter(0);
        string eventName = GetParameter(1);
        string eventType = GetParameter(2);
        string eventCategory = GetParameter(3);

        if (functionName == "createEvent")
            eventController.CreateEvent (eventName, eventType, eventCategory);
    }
}
```

Figure 5.10 SequencerCommand *GenerateEvent*

Thus, using the mechanism explained above, a new event can be created and added to the event log when the student refuses to teach the water molecule. Other dialogue events are generated in a similar way.

**Time Events**. In VS Saga, only one time event is tracked. The Event Control component keeps a timer for the student's inactive time. Inactive time is defined as the time when the student is not actively interacting with the game characters. Whenever a dialogue event is generated, the timer will be reset. When the timer times out, a time-out event will be generated which signals that the student has been inactive.

**Teaching Feedback Events**. A teaching feedback event is an event generated to provide feedback to the student's teaching. Two such events are tracked in VS Saga, namely Teach Success Event and Teach Failure Event.

The list of events that are tracked in VS Saga are summarized in **Appendix 3**.

**5.4.2 FCM Calculation**

Instead of using existing FCM tools, a FCM Calculation component is developed so that it can be easily integrated with other control components developed in Unity. The FCM Calculation component is implemented according to the FCM model proposed in section 3.3.2. The complete PTA FCM used in VS Saga can be found in **Appendix 4**.

A class diagram of the FCM Calculation component is shown in Figure 5.11. From the bottom of the diagram, the *AdjacencyList* class stores the adjacency list representation of the PTA FCM. The PTA FCM can be represented by a sparse matrix. Compared to a 2D matrix, adjacency list is a more suitable data structure to model sparse matrix as it has higher space efficiency.

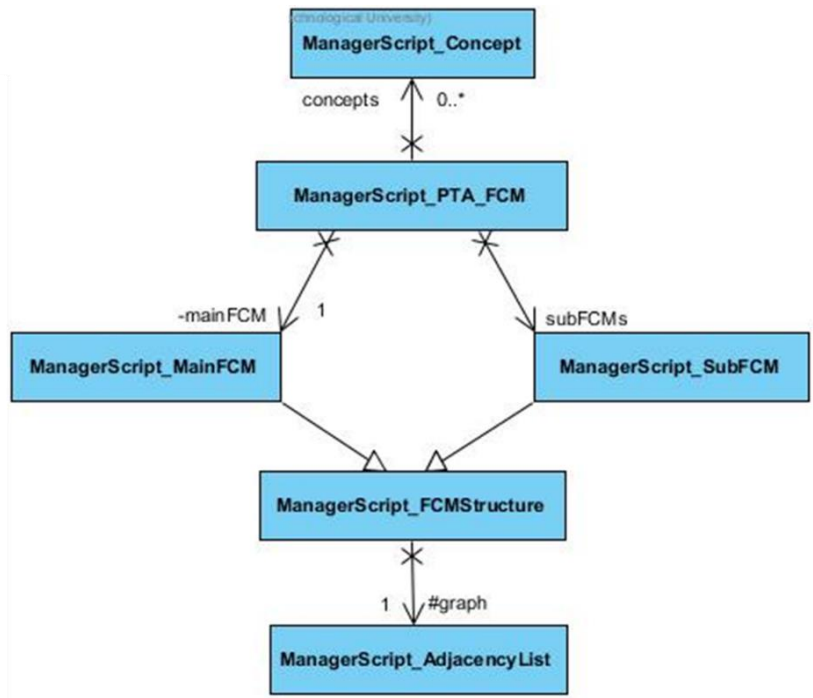

Figure 5.11 Class Diagram of FCM Calculation Component

The class *FCMStructure* has an instance of *AdjacencyList* class and contains the logic for the computational processes of FCM. It provides a method called *matrixCalculation()* which can take in the values of the all causal concepts, perform one round of calculation and return the updated values of the causal concepts.

*MainFCM* and *SubFCM* are two subclasses of *FCMStructure* and inherit all its attributes and methods. The concept of mainFCM and subFCM are defined in **Definition 7** and **Definition 8**. The *SubFCM* class overrides the *matrixCalculation()* method in *FCMStructrue* as it uses a slightly different way to perform its computational processes.

The *PTA_FCM* class models the structure and characteristics of a PTA FCM. It has an instance of *MainFCM* class and multiple instances of *SubFCM* class. The *PTA_FCM* class also has a list of *Concepts* which is used to store the values of the causal concepts. The *PTA_FCM* class coordinates the calculation processes in the mainFCM and multiple subFCMs. Multiple rounds of calculations maybe needed before the values of causal concepts finally stabilize.

This implementation of the FCM Calculation component caters to the structure of PTA FCM by modelling mainFCM and subFCMs separately and differentiates their computational processes. In some cases, separating the computations in mainFCM and subFCMs can also increase the overall computational efficiency. For example, for a PTA FCM which uses a trivalent threshold function as shown in Figure 5.12, starting from the second round of calculation, computations only need to be carried out in the mainFCM. By reducing the amount of computations performed, the efficiency of the overall processes is improved.

$$f(x) = \begin{cases} -1 & x \leq -0.5 \\ 0 & -0.5 < x < 0.5 \\ 1 & 0.5 \leq x \end{cases}$$

Figure 5.12 Trivalent Threshold Function

### 5.4.3 UI Control

The UI Control component is implemented as a set of scripts which controls different UIs. This section provides a few examples to show how the UI Control manages the different UIs.

**Persuasion Reasoning**. The TA panel (Figure 5.13) is the most important interface element for the Persuasion Reasoning. It is used for displaying the persuasion cues. After a persuasion cue is selected, the PTA Control component calls the UI Control to fetch the information of the selected cue and display it. The information to be fetched from the database includes the texts to be displayed and the facial expression of the water molecule.

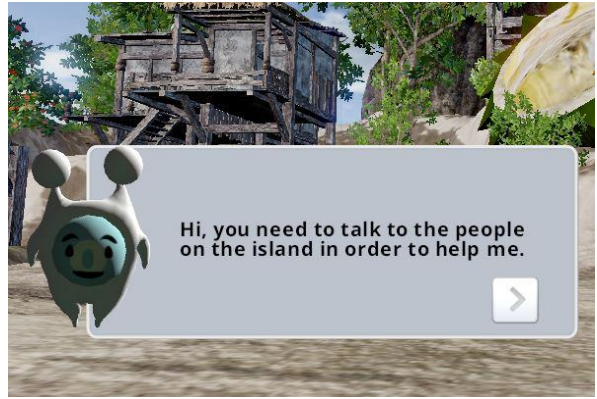

Figure 5.13 TA Panel

**Teachability Reasoning**. The concept map (Figure 5.14) is the main interface where the teaching process is performed. In order to teach the PTA (the water molecule), the student completes the concept map by dragging and dropping the buttons on the right side to the blank spaces on the left. When the "Teach" button is clicked, the UI Control saves the contents of the blank spaces as the knowledge learnt and creates a new event to indicate the completion of the teaching process.

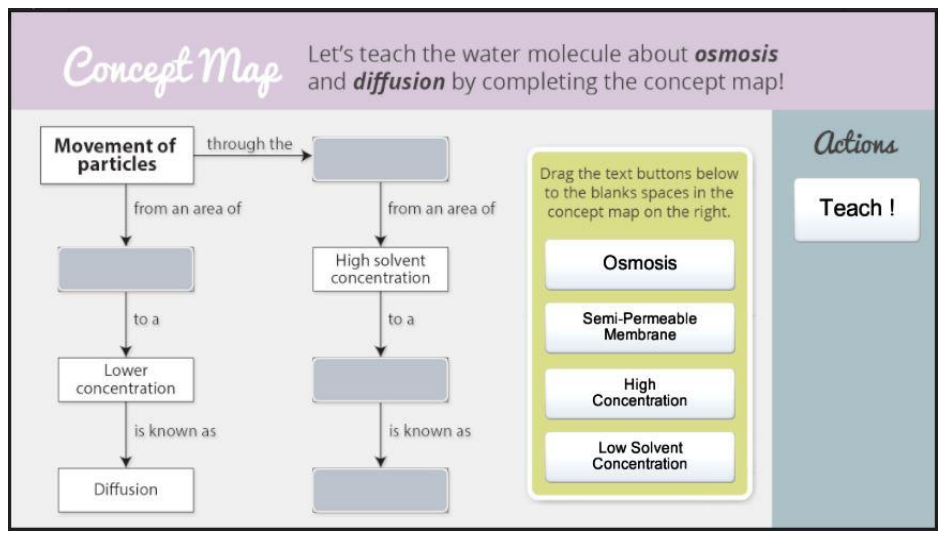

Figure 5.14 Concept Map

**Practicability Reasoning**. The banana tree is the most important interface element for the Practicability Reasoning. If the student teaches the PTA correctly, the UI Control will change the color and positions of the tree leaves to make the tree become revitalized. If the student commits errors during teaching, the tree remains withered.

## 5.5 Database Access

In VS Saga, the Data Access component needs to provide interfaces for two database management systems, MySQL and SQLite. The Goal Net database and FCM database are managed by MySQL while the database for persuasion cues and inventory items is managed by SQLite.

Inventory information and persuasion cues are stored in the same database which is managed by SQLite. Thus, there are no access issues. However, the Goal Net database and FCM database are two separate databases and both managed by MySQL. Since the best practice is to keep only one open connection at any time, accesses to these two databases need to be coordinated. The Database Access loads the FCM model from the FCM database during initialization and closes the connection to the FCM database. Then, it opens the connection to the Goal Net database and executes queries while the Goal Net Interpreter is traversing the Goal Net of the PTA.

# Chapter 6 Case Studies and Assessment of PTA

## 6.1 Case Studies of PTA

This section presents a few case studies of the implemented PTA. It aims to demonstrate that the PTA implemented in VS Saga has realized all major characteristics of a PTA. These case studies are going to be used in future submission for the **IJCAI-15 Video Competition**.

### 6.1.1 Persuasion Case Studies

**Case Study 1: Not Learning Diffusion**

1.1 Meet Sharman near the big durian in the Knowledge Town

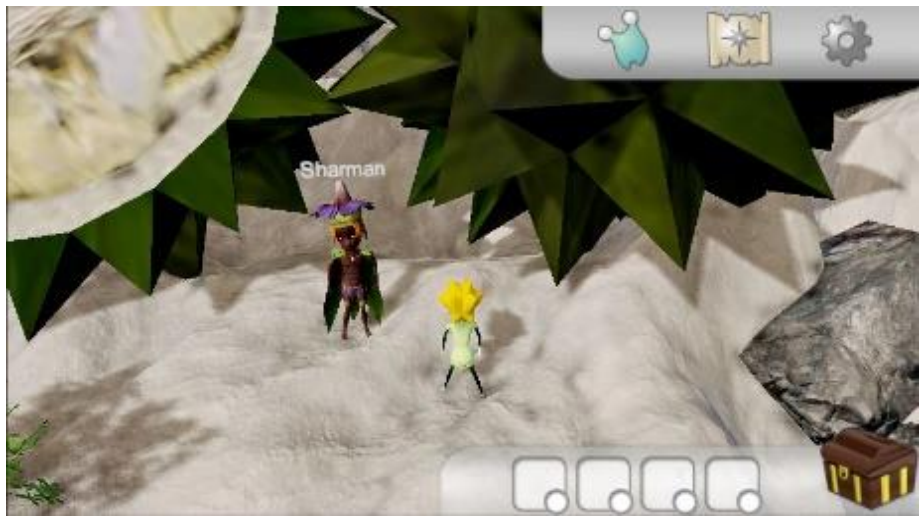

Figure 6.1(a) Meet Sharman

1.2 Talk to Sharman. Sharman mentions about diffusion.

    1.2.1 Choose the response "A. Diffuse? What do you mean by diffuse?"

    => High motivation, high ability state

    1.2.2 Choose the response "B. Ok…. I see…."

    => Low motivation, low ability state, **go to 1.3**

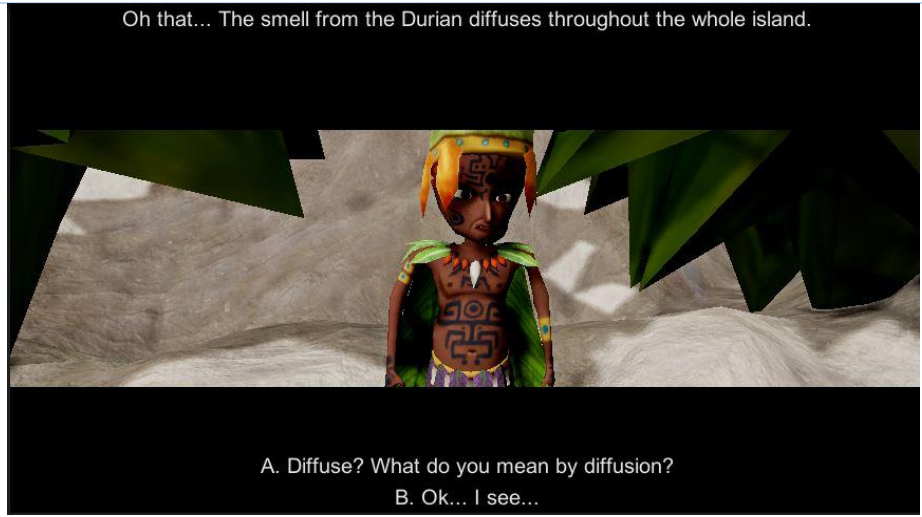

Figure 6.1(b) Talk to Sharman

1.3 The water molecule (the PTA) persuades the student to learn about diffusion

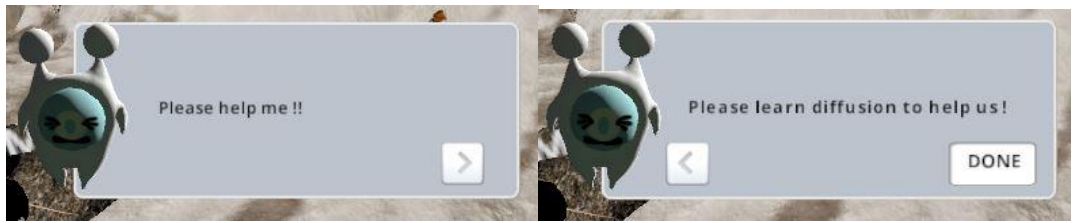

Figure 6.1(c) Persuasion for Not Learning Diffusion

**Case Study 2: Not Motivated to Conduct Experiments**

2.1 Meet the Future Teacher in the Science Laboratory

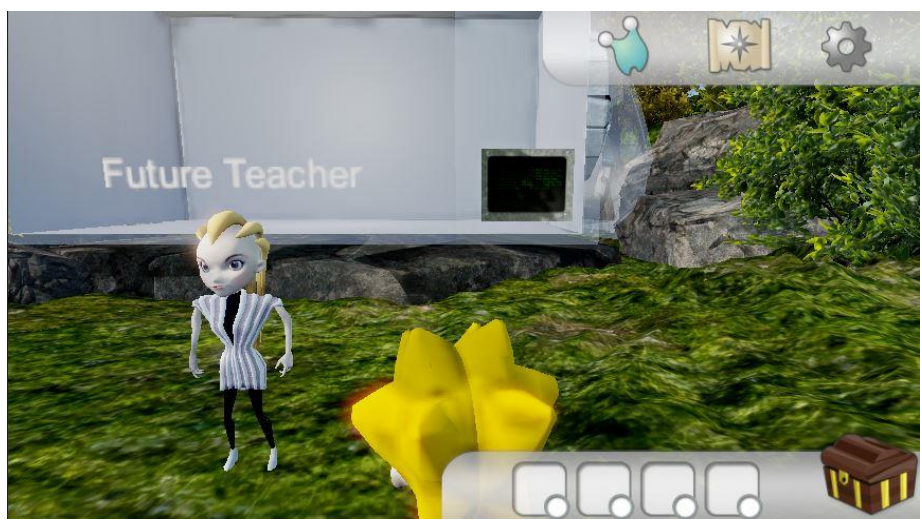

Figure 6.2(a) Meet the Future Teacher

2.2 The Future Teacher mentions about "Diffusion 5K", the big tank for conducting diffusion and osmosis experiments.

    2.2.1 Choose the response "Ok. But how should I operate this Diffusion 5K?"

    => High motivation high ability state

    2.2.2 Choose the response "B. Ok. Probably later."

    => Low motivation low ability state, **go to 2.3**

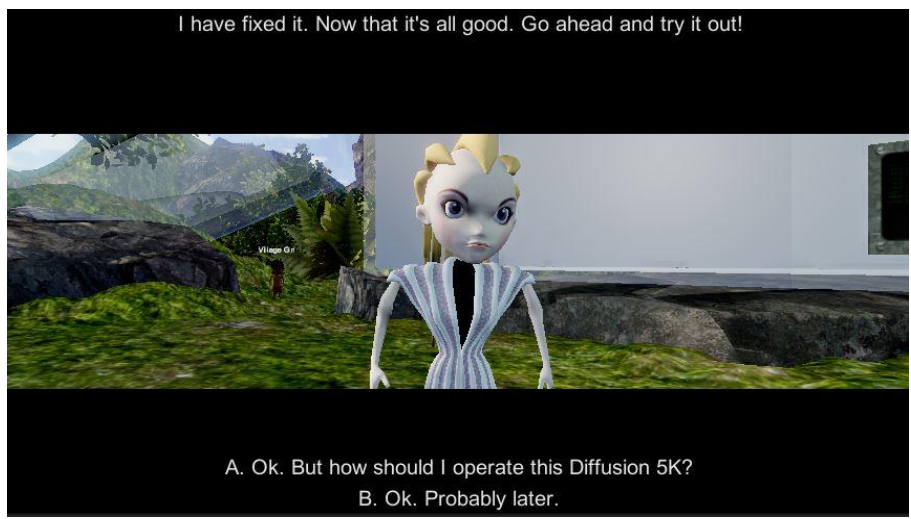

Figure 6.2(b) Talk about Diffusion 5K

2.3 The water molecule persuades the student to try out the experiments.

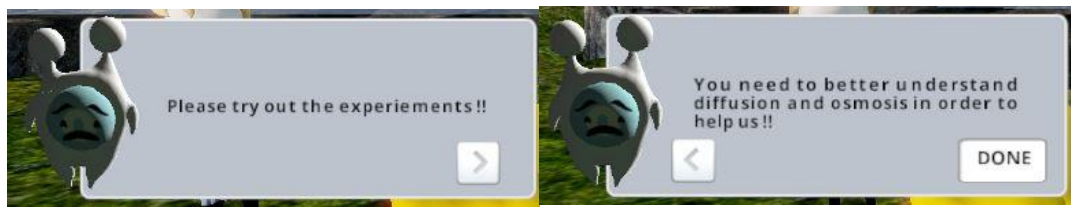

Figure 6.2(c) Persuasion for Not Motivated to Conduct Experiments

**Case Study 3: Talking to Distracting Characters**

3.1 Meet the Rabbit near the stairs in the Knowledge Town

    3.1.1 Choose not to talk to the rabbit

    => High motivation high ability state

    3.1.2 Choose to talk to the rabbit, **go to 3.2**

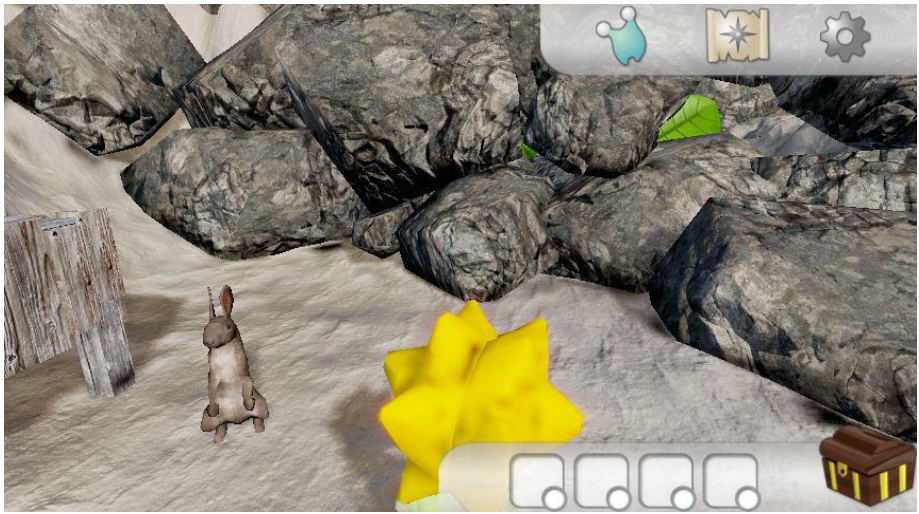

Figure 6.3(a) Meet the Rabbit

3.2 Have conversation with the rabbit

    => Low motivation low ability state, **go to 3.3**

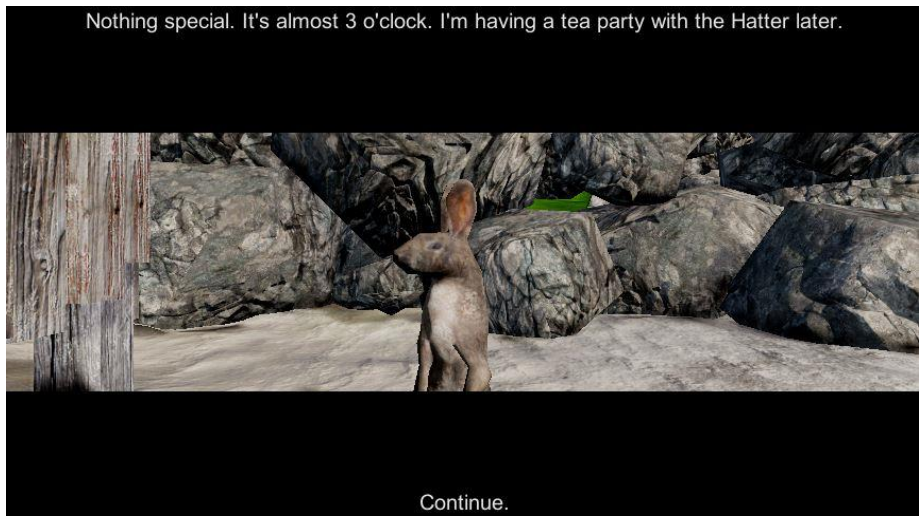

Figure 6.3(b) Talk to the Rabbit

3.3 The water molecule (the PTA) persuades the student to concentrate on his or her mission and do not play around.

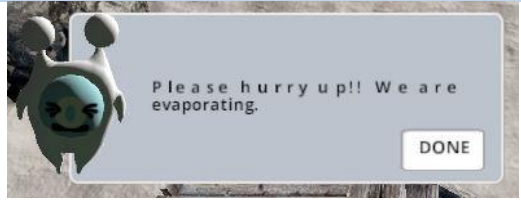

Figure 6.3(c) Persuasion for Talking to Distracting Characters

### 6.1.2 Teachability and Practicability Case Studies

**Case Study 4: Teach Failure**

4.1 Arrive at the Banana Tree. See the sad water molecule and the withered tree

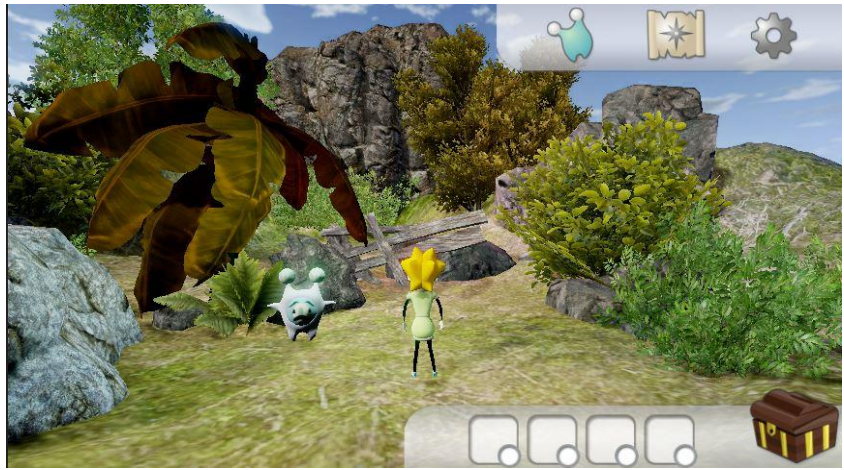

Figure 6.4(a) Arrive at the Banana Tree

4.2 Commit errors while drawing the concept map

=> Click the "Teach!" button, **go to 4.3**

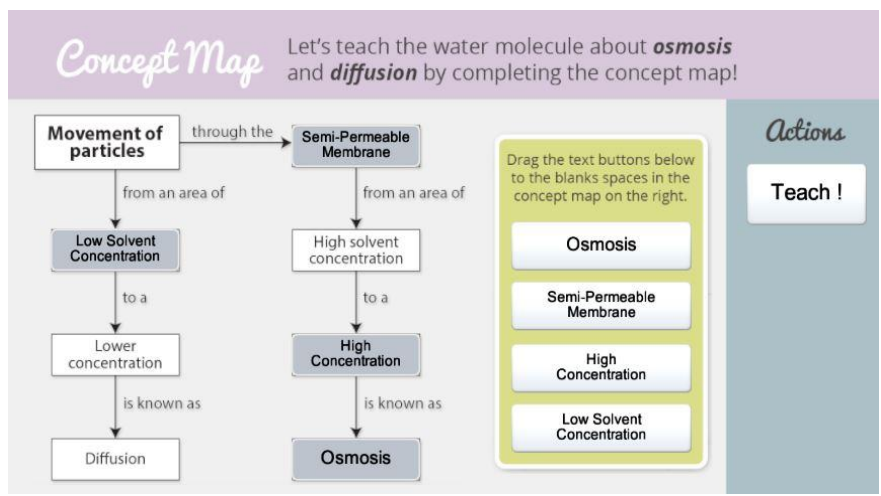

Figure 6.4(b) Commit Errors during Teaching

4.3 After the Practicability Reasoning, no solution can be generated. Thus, the Persuasion Reasoning calls up the water molecule to persuade the student to teach it again.

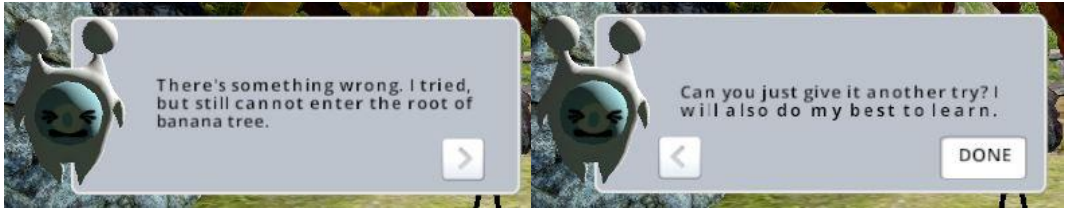

Figure 6.4(c) Persuasion for Teaching Failure

**Case Study 5: Teach Success after Previous Failure**

5.1 Teach the water molecule again after fail for the first time. The errors committed during previous teaching process are highlighted in red.

    5.1.1 Fail to correct all the errors

    => Click on "Teach!, **go to 4.3**

    5.1.2 Correct all the errors

    => Click on "Teach!", **go to 5.2**

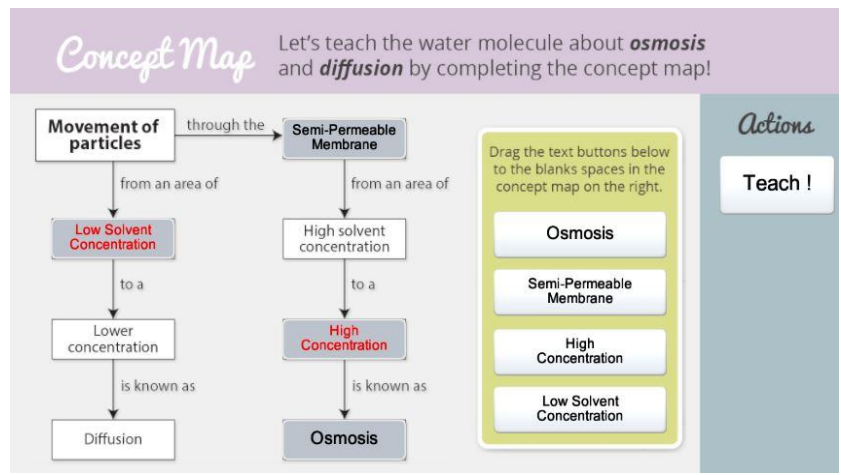

Figure 6.5(a) Teach the Water Molecule Again

5.2 After the Practicability Reasoning, a correct solution is derived. The Practicability Reasoning calls up the UI Control module of the tree to display the tree animation. The tree becomes revitalized and the water molecule becomes happy.

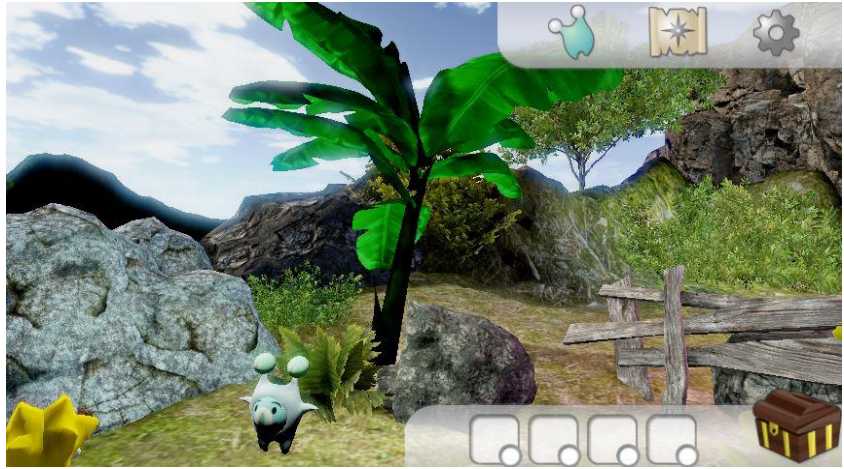

Figure 6.5(b) Revitalized Tree and Happy Water Molecule

5.3 The water molecule gives positive feedback on the teaching.

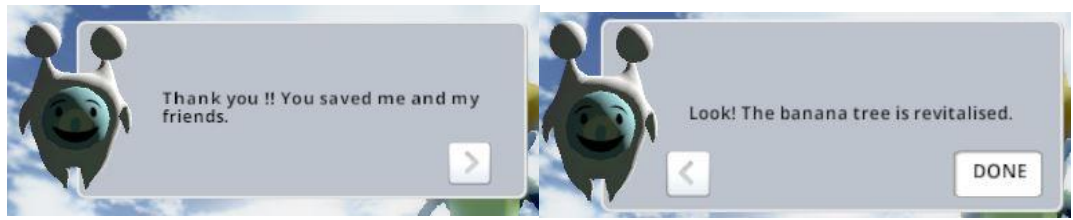

Figure 6.5(c) Happy Molecule and Mission Accomplished

## 6.2 Qualitative Assessment of PTA

### 6.2.1 Assessment Approach – Focus Group Study

For the qualitative assessment of the PTA, a focus group study is carried out in order to gather feedback, comments and views on the PTA. Since the PTA sits in the interdisciplinary research field of education, intelligent agent and psychology, the focus group should include participants from various fields in order to get feedback and comments on various aspects of the PTA. The design of the focus group study is summarized in Table 6.1.

Table 6.1 The Design of the Focus Group Study

| Design of the Focus Group Study | |
|---|---|
| Number of Participants | 6 |
| Formation of the Participants | 2 Secondary school teachers<br>2 Agent researchers<br>2 Game developers |
| Focus Group Main Questions | A. Regarding the effectiveness of the PTA as a learning companion:<br>　(1) Do you think the PTA is compelling enough to affect the minds and decisions of the student during the learning process and why?<br>　(2) How do you think the PTA can help the student to grasp better understanding of the learning topics and why?<br>　(3) How do you think the PTA is able to improve the learning outcomes of the student and why?<br>　(4) How do you think the PTA can affect the student's attitudes towards learning?<br>　(5) Any areas for improvements?<br>B. Regarding the agent model of the PTA:<br>　(1) Do you think the proposed agent model would increase the spontaneity of the agent and why?<br>　(2) Do you think the proposed agent model would help the PTA to generate personalized feedback and why?<br>　(3) Is the modelling of the agent considered complete and flexible and why?<br>　(4) Any areas for improvements?<br>C. Regarding the system design and implementation of the |
| | PTA: <br> (1) Is the system design reasonable? <br> (2) Based on the proposed agent model, how would you design the system architecture? <br> (3) Do you have any comment regarding the maintainability and reusability of the system? <br> (4) Any areas for improvements? |
| Activity Flow | A. a short introduction a the PTA model and system architecture (15 min) <br> B. a short demo of the VS Saga game (5 min) <br> C. Moderated focus group discussion (45 min) <br> D. A short interview with each participant afterwards (10 min each) |

### 6.2.2 Result Analysis of the Focus Group Study

As the number of participants in the focus group study is small, no statistical significance analysis can be drawn from the study. However, it is valuable to summarize all the qualitative assessments to identify the advantages and the areas for improvements of the PTA. The feedback and comments gathered from the study are summarized in the following table.

Table 6.2 The Results from the Focus Group Study

| | Advantages |
|---|---|
| Teachers | ▪ The water molecule has human-like emotions and is proactive in interactions with student. The student can develop sympathy and sense of responsibility towards the agent which motivates them to learn himself or herself and teach the agent. <br> ▪ The student is unconsciously "supervised" by the water molecule. He or she can receive individual attention of the water molecule and be persuaded immediately if he or she is demotivated. |
|  | ▪ The teaching process can help to reinforce learning of the student. |
|  | ▪ The learning and teaching processes are designed as a more enjoyable experience than the traditional class room style. The processes would be engaging. |
|  | Areas for Improvement |
|  | ▪ The game and the PTA should be able to cater to different learning contents, not only for diffusion and osmosis. |
|  | ▪ When evaluating the ability level of the student, his or her ability to learn and prior knowledge should also be taken into consideration. If the student is slow learner and has little prior knowledge in related topics, he or she should be allowed more time to learn instead of being persuaded immediately. |
| **Agent Designers** | Advantages |
|  | ▪ The goal-oriented approach enables the agent to direct its own behavior and perform tasks in the pursuit of its own goal. |
|  | ▪ The propose Goal Net Model describes a PTA in all its aspect and accommodates flexibility. |
|  | ▪ The integration of FCM enhances the qualitative reasoning ability of the agent. |
|  | Areas for Improvement |
|  | ▪ The agent is implemented specifically for topics in the science domain. However, the design of the agent should incorporate the flexibility so it can be deployed for more topics and in more domains. |
| **Game Developers** | Advantages |
|  | ▪ The system architecture is easy to understand. It's abstract enough to be reusable. |
|  | ▪ I would implement it in similar ways. |
|  | Areas for Improvement |
|  | ▪ Currently, the task functions are implemented in Unity. They can |

> - also be implemented as DLL files, which can be more easily called by a virtual machine (the interpreter).
> - Most teachers are computer novices who are not familiar with game development and agent design. Convenient update tools should be developed for them, if the teachers need to update the learning content frequently.

Generally, the results of the assessment are quite positive. The results show that the PTA has the ability to motivate and engage the student by generating personalized feedback and the potential to help the student build positive attitude towards learning. It also turns the learning process into an enjoyable journey which could potentially help the student build positive attitudes towards learning. In terms of modelling, the Goal Net model depicts a PTA in all its aspects. The system architecture has also been endorsed for its reasonableness and reusability.

Another important advantage of the PTA has been put forward during the focus group discussion. Compared to the human teacher, the PTA can generate more timely feedback, as it is able to closely "supervise" the student by monitoring his or her motivation and ability level using a more direct and quantitative approach.

### 6.2.3 Limitations and Areas for Improvements

The focus group study has also highlighted several limitations and areas for improvement. The one that has been raised by most of the focus group participants is the lack of flexibility of the current PTA. The instance implemented in VS Saga is specifically for topics in the science domain. Greater flexibility needs to be incorporated so that the PTA and the pedagogical game can be customized to different learning topics. In order to do this, there is a need to propose a convenient way to modify the models of the agent and change the content of the knowledge base. One way to incorporate that kind of flexibility is to design an authoring tool for the PTA, so that the design and implementation process of the PTA can be more smoothly connected. This is an area which the future researches in PTA can look into. Another important area for improvement is to consider the learning competency and prior knowledge of the student while assessing his or her ability level.

# Chapter 7 Conclusion and Future Work

## 7.1 Conclusions

A Persuasive Teachable Agent (PTA) is a special type of Teachable Agent (TA) which has the ability to be taught, to practice knowledge learnt and to persuade the student to teach it. By incorporating the Elaboration Likelihood Model (ELM) of persuasion, the PTA is able to provide more timely and personalized feedback based on the student's ability and motivation, which previous TAs are incapable of.

However, the existing PTA model still has a few limitations. Firstly, the existing agent model has only proposed the Goal Net model for the Persuasion Reasoning while does not have Goal Net models for the Teachability Reasoning and the Practicability Reasoning. Secondly, the FCM model for the Persuasion Reasoning is difficult to reuse as it is highly context dependent. Thirdly, there is still a gap between theoretical models and practical implementation of the PTA.

In this project, a complete and integrate PTA agent model was proposed. Directed by a goal-oriented approach, the agent model depicts a PTA in its totality, including its ability to perform Persuasion Reasoning, Teachability Reasoning and Practicability Reasoning. This project also proposed a FCM model with greater reusability for computational processes in the Persuasion Reasoning of the PTA. In some cases, the proposed FCM model is able to show improved overall computational efficiency. Besides proposing improvements for the theoretical models of the PTA, system architecture for the control structure of the agent was introduced as well, with detailed descriptions for each component in the architecture and how it can be implemented.

Following the proposed agent model and system architecture, an instance of the PTA has been successfully deployed in the 3D videogame, VS Saga. This implementation demonstrated the practice of instantiating a PTA from the propose agent model and system architecture. It also enabled the assessment and evaluation of the PTA. A focus group study was performed to assess the PTA.

The results of the study demonstrated the effectiveness of the PTA. The results were quite positive, showing the PTA is able to motivate and engage the student by generating timely and personalized feedback. It was also believed that the PTA has the potential to help the student build positive attitude towards learning. Moreover, positive feedback has been received for the completeness of the new agent model and the reusability of the proposed system architecture.

## 7.2 Future Work

### 7.2.1 Quantitative Assessment of PTA

During the focus group study, no statistically significant analysis can be drawn from the results since the number of participants is too small. No conclusions can be drawn regarding how the PTA will affect the learning outcomes of the student as well.

Thus, quantitative assessment of the PTA has to be carried out in order to analyze the effectiveness of the PTA statistically. VS Saga can be deployed onto different platforms for the assessment purpose, for example, desktops and tablets. The assessment could be conducted with a pre-test, a post-test and a set of questionnaire. The pre-test and the post-test can contain questions to assess the student's understanding of the content knowledge. By comparing the results of the two tests, the student's knowledge gain through teaching the PTA and his or her learning outcomes can be assessed. The questionnaire can be used to complement the tests and survey the attitudes of the students and their feelings towards the PTA.

### 7.2.2 PTA Authoring Tool

As discussed in section 6.2.3, the results of the focus group study have pointed out a few directions for future research. Before the PTA can be widely deployed, it needs to build up the flexibility to cater to different learning topics.

For different learning topics, the learning goal and domain knowledge involved would be different. Thus, there is a need to propose a convenient way to modify the models of the agent and change the content of the knowledge base. One way to incorporate that kind of flexibility is to design an authoring tool for the PTA. Referring to the design of the authoring tool which Ailiya proposed for the Affective Teachable Agent [11], a similar design can be proposed for the authoring tool of the PTA (Figure 7.1). A dedicated Game Authoring Component can be created, which provides interfaces for creating and editing learning goals, content knowledge and game tasks. If any changes are made from the Game Authoring Component, the models and knowledge stored in the knowledge base should be updated accordingly.

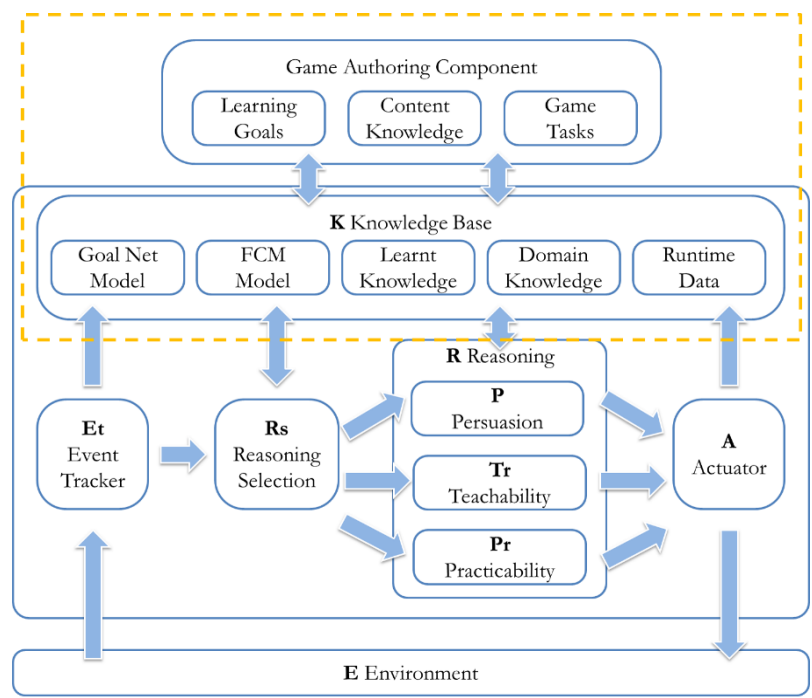

Figure 7.1 The Proposed Authoring Component of PTA

Authoring processes would involve both the teacher and the game developer. The teacher defines the learning goals and relevant content knowledge. The teacher then designs different game tasks according to the learning goals. According to the teacher's design, the game developer modifies the game and add create new game tasks.

### 7.2.3 Application of Educational Data Mining

One big advantage of educational games is providing a digital ambience in which a great amount of behavioral data of the student can be collected. These data can be tracked and analyzed to generate useful feedback to the student concerning his or her learning.

Table 4.3 lists 7 types of event data that can be tracked in the games. However, in VS Saga, the assessment of motivation and ability level is mostly based on instant event data and only 3 out of these 7 types of events are tracked.

In future studies, efforts can be directed to increase the type of events tracked. By analyzing more sources of events, the PTA will be able to assess the motivation and ability level of the student more accurately. Moreover, future researches could look into the application of educational data mining to analyze the relationship between different types of student behavioral data and the level of motivation and ability.

For example, as suggested in Table 4.3, Location Events can be tracked in the PTA environment. By collecting the location of students in the virtual environment every few seconds, abundant location data can be obtained. Data mining techniques can be applied to these location data to mine the location patterns of the student. Relationships can be established between these patterns and the motivation and ability level of the student. For example, if the points indicating locations are clustered around the main characters in the scenes and the banana tree with only a few outliers, the student has probably been focusing on his or her mission on the island and is potentially in a high motivation state.

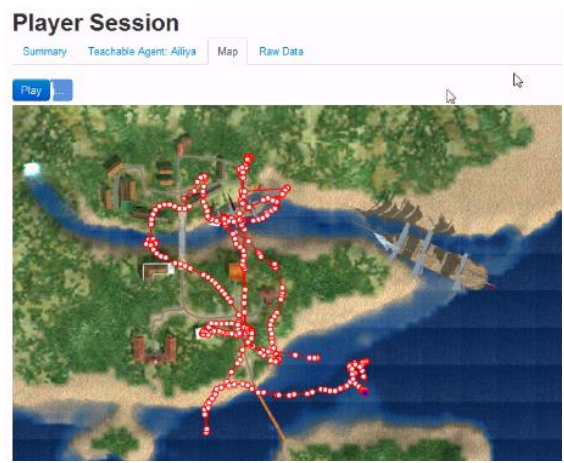

Figure 7.2 Tracking the Location Events

# References

[1] G. K. D. N. Biswas, "LEARNING BY TEACHING: A NEW AGENT PARADIGM FOR EDUCATIONAL SOFTWARE," *Applied Artificial Intelligence,* vol. 19, no. 3/4, pp. 363-392, 03//, 2005.

[2] S. F. Lim, Ailiya, C. Miao, and Z. Shen, "The Design of Persuasive Teachable Agent.", in *Advanced Learning Technologies, 2013 IEEE 13th International Conference, ICALT 2013, 15-18 July 2013, Beijing, China.* pp. 382-384. Available: IEEE Xplore, http://www.ieee.org.

[3] K. g. b. v. e. Leelawong, and G. g. b. v. e. Biswas, "Designing Learning by Teaching Agents: The Betty's Brain System," *International Journal of Artificial Intelligence in Education (IOS Press),* vol. 18, no. 3, pp. 181-208, 10//, 2008.

[4] N. Matsuda, E. Yarzebinski, V. Keiser, R. Raizada, W. W. Cohen, G. J. Stylianides, and K. R. Koedinger, "Cognitive anatomy of tutor learning: Lessons learned with SimStudent," *Journal of Educational Psychology,* vol. 105, no. 4, pp. 1152-1163, 2013.

[5] Ailiya, Z. Shen, and C. Miao, "Affective Teachable Agent in VLE: A Goal Oriented Approach.", in *Advanced Learning Technologies, 2011 IEEE 11th International Conference, ICALT 2011, 6-8 July 2011, Athens, GA.* pp. 110-114. Available: IEEE Xplore, http://www.ieee.org.

[6] M. Leiba, R. Zuzovsky, D. Mioduser, Y. Benayahu, and R. Nachmias, "Learning about Ecological Systems by Constructing Qualitative Models with DynaLearn," *Interdisciplinary Journal of E-Learning & Learning Objects,* vol. 8, pp. 165-178, 2012.

[7] K. Kapur, *Building learning mini-games*, 2011. Submitted to to the School of Computer Engineering of the Nanyang Technological University

[8] K. Squire, and S. Barab, "Replaying history: engaging urban underserved students in learning world history through computer simulation games." in *Proceedings of the 6th international conference on Learning Sciences, ICLS 2004,* pp. 505-512. Available: ACM Digital Library, http://dl.acm.org.

[9]     M. de Aguilera, and A. Mendiz, "Video games and education," *Computers in Entertainment,* vol. 1, no. 1, pp. 1, 10//, 2003.

[10]    K. Blair, D. Schwartz, G. Biswas, and K. Leelawong, "Pedagogical agents for learning by teaching: Teachable agents," *EDUCATIONAL TECHNOLOGY-SADDLE BROOK THEN ENGLEWOOD CLIFFS NJ-,* vol. 47, no. 1, pp. 56, 2007.

[11]    A. Borjigin, *Affective teachable agent in virtual learning environment*, 2014. Submitted to to the School of Computer Engineering of the Nanyang Technological University

[12]    J. Gaustad, and E. O. R. Eric Clearinghouse on Educational Management, *Peer and Cross-Age Tutoring. ERIC Digest, Number 79*, 1993.

[13]    R. H. Gass, and J. S. Seiter, *Persuasion, social influence, and compliance gaining*: Boston : Allyn & Bacon, c2011. 4th ed., 2011.

[14]    R. E. Petty, and J. T. Cacioppo, "The Elaboration Likelihood Model of Persuasion," *Advances in Experimental Social Psychology,* vol. 19, pp. 123-205, 1/1/1986, 1986.

[15]    B. Kosko, "Fuzzy cognitive maps," *International Journal of Man-Machine Studies,* vol. 24, pp. 65-75, 1/1/1986, 1986.

[16]    Z. Shen, *Goal-oriented modeling for intelligent agents and their applications*, 2005. Submitted to to the School of Computer Engineering of the Nanyang Technological University

[17]    C. Dede, J. Clarke, D. J. Ketelhut, B. Nelson, and C. Bowman, "Students' motivation and learning of science in a multi-user virtual environment." in *American Educational Research Association Conference 2005*, *Montreal, Canada*.

[18]    H. Yu, Z. Shen, and C. Miao, "Intelligent Software Agent Design Tool Using Goal Net Methodology." in *Intelligent Agent Technology 2007, IAT '07. IEEE/WIC/ACM International Conference*, *2-5 November 2007*, *Fremont, CA*. pp. 43-46. Available: IEEE Xplore, http://www.ieee.org.

[19]    J. D. Novak, "Concept mapping: A useful tool for science education," *Journal of research in science teaching,* vol. 27, no. 10, pp. 937-949, 1990.

[20] S. F. Lim, "Persuasive Teachable Agent for Intergenerational Learning," 2014. Submitting to to the School of Computer Engineering of the Nanyang Technological University

[21] H. Yu, Z. Shen, and C. Miao, "A goal-oriented development tool to automate the incorporation of intelligent agents into interactive digital media applications," Computers in Entertainment (CIE), vol. 6, no. 2, p. 24, 2008.

[22] ——, "Intelligent software agent design tool using goal net methodology," in Proceedings of the 2007 IEEE/WIC/ACM International Conference on Intelligent Agent Technology. IEEE Computer Society, 2007, pp. 43–46.

[23] H. Yu, Y. Cai, Z. Shen, X. Tao, and C. Miao, "Agents as intelligent user interfaces for the net generation," in Proceedings of the 15th international conference on Intelligent user interfaces. ACM, 2010, pp. 429–430.

[24] B. Li, H. Yu, Z. Shen, and C. Miao, "Evolutionary organizational search," in Proceedings of The 8th International Conference on Autonomous Agents and Multiagent Systems-Volume 2. International Foundation for Autonomous Agents and Multiagent Systems, 2009, pp. 1329–1330.

[25] H. Yu, C. Miao, X. Tao, Z. Shen, Y. Cai, B. Li, and Y. Miao, "Teachable agents in virtual learning environments: a case study," in World Conference on E-Learning in Corporate, Government, Healthcare, and Higher Education, vol. 2009, no. 1, 2009, pp. 1088–1096.

[26] H. Yu, Z. Shen, and C. Miao, "A trustworthy beacon-based location tracking model for body area sensor networks in m-health," in Information, Communications and Signal Processing, 2009. ICICS 2009. 7th International Conference on. IEEE, 2009, pp. 1–5.

[27] H. Yu, Z. Shen, C. Miao, and A.-H. Tan, "A simple curious agent to help people be curious," in The 10th International Conference on Autonomous Agents and Multiagent Systems-Volume 3. International Foundation for Autonomous Agents and Multiagent Systems, 2011, pp.

1159–1160.

[28] [63] C. Miao, H. Yu, Z. Shen, and X. Tao, "Agents for collaborative learning in virtual worlds," Workshop on Opportunities for intelligent and adaptive behavior in collaborative learning systems, p. 21, 2010.

[29] J. Lin, C. Miao, and H. Yu, "A cloud and agent based architecture design for an educational mobile sns game," in Edutainment Technologies. Educational Games and Virtual Reality/Augmented Reality Applications. Springer Berlin Heidelberg, 2011, pp. 212–219.

[30] X. Tao, Z. Shen, C. Miao, Y.-L. Theng, Y. Miao, and H. Yu, "Automated negotiation through a cooperative-competitive model," in Innovations in Agent-Based Complex Automated Negotiations. Springer Berlin Heidelberg, 2011, pp. 161–178.

[31] H. Yu, S. Liu, A. C. Kot, C. Miao, and C. Leung, "Dynamic witness selection for trustworthy distributed cooperative sensing in cognitive radio networks," in Proceedings of the 13th IEEE International Conference on Communication Technology (ICCT'11). IEEE, 2011, pp. 1–6.

[32] Z. Shen, H. Yu, C. Miao, and J. Weng, "Trust-based web service selection in virtual communities," Web Intelligence and Agent Systems, vol. 9, no. 3, pp. 227–238, 2011.

[33] Y.-L. Theng, K.-L. Tan, E.-P. Lim, J. Zhang, D. H.-L. Goh, K. Chatterjea, C. H. Chang, A. Sun, H. Yu, N. H. Dang et al., "Mobile g-portal supporting collaborative sharing and learning in geography fieldwork: an empirical study," in Proceedings of the 7th ACM/IEEE-CS joint conference on Digital libraries. ACM, 2007, pp. 462–471.

[34] H. Yu, Z. Shen, and C. Leung, "Towards trust-aware health monitoring body area sensor networks," International Journal of Information Technology, vol. 16, no. 2, 2010.

[35] L. Pan, X. Meng, Z. Shen, and H. Yu, "A reputation-based trust aware web service interaction pattern for manufacturing grids," International Journal of Information Technology (IJIT), vol. 17, no. 1, 2011.

[36] H. Yu, Z. Shen, C. Miao, C. Leung, and D. Niyato, "A survey of trust and reputation management systems in wireless communications," Proceedings of the IEEE, vol. 98, no. 10, pp. 1755–1772, 2010.

[37] T. Qin, H. Yu, C. Leung, Z. Shen, and C. Miao, "Towards a trust aware cognitive radio architecture," ACM SIGMOBILE Mobile Computing and Communications Review, vol. 13, no. 2, pp. 86–95, 2009.

[38] L. Pan, X. Meng, Z. Shen, and H. Yu, "A reputation pattern for service oriented computing," in Information, Communications and Signal Processing, 2009. ICICS 2009. 7th International Conference on. IEEE, 2009, pp. 1–5.

[39] H. Yu, Z. Shen, C. Miao, J. Wen, and Q. Yang, "A service based multi-agent system design tool for modelling integrated manufacturing and service systems," in Emerging Technologies and Factory Automation, 2007. ETFA. IEEE Conference on. IEEE, 2007, pp. 149–154.

[40] H. Yu, Z. Shen, and B. An, "An adaptive witness selection method for reputation-based trust models," PRIMA 2012: Principles and Practice of Multi-Agent Systems, pp. 184–198, 2012.

[41] H. Yu, C. Miao, X. Weng, and C. Leung, "A simple, general and robust trust agent to help elderly select online services," in Network of Ergonomics Societies Conference (SEANES), 2012 Southeast Asian. IEEE, 2012, pp. 1–5.

[42] C. Leung, C. Miao, H. Yu, and M. Helander, "Towards an ageless computing ecosystem," International Journal of Information Technology (IJIT), vol. 18, no. 1, 2012.

[43] P. Cheng, H. Yu, Z. Shen, and Z. Liu, "An interactive 3d product design tool for mobile pre-commerce environments," International Journal of Information Technology (IJIT), vol. 18, no. 2, 2012.

[44] H. Yu, Z. Shen, C. Miao, and B. An, "Challenges and opportunities for trust management in crowdsourcing," in IEEE/WIC/ACM International Conference on Intelligent Agent Technology (IAT). IEEE, 2012, pp. 486–

493.

[45] ——, "A reputation-aware decision-making approach for improving the efficiency of crowdsourcing systems," in The 12th International Conference on Autonomous Agents and Multi-Agent Systems (AAMAS'13), 2013.

[46] S. Liu, H. Yu, C. Miao, and A. C. Kot, "A fuzzy logic based reputation model against unfair ratings," in The 12th International Conference on Autonomous Agents and Multi-Agent Systems (AAMAS'13), 2013.

[47] H. Yu, Z. Shen, C. Leung, C. Miao, and V. R. Lesser, "A survey of multi-agent trust management systems," IEEE Access, vol. 1, no. 1, pp. 35–50, 2013.

[48] H. Yu, Y. Cai, Z. Shen, X. Tao, and C. Miao, "Intelligent learning companions for virtual learning environments," in Multi-agent in Education and Entertainment Workshop in the 9th International Conference on Autonomous Agents and Multi-agent Systems (AAMAS'10), 2010.

[49] H. Song, Z. Shen, H. Yu, and Y. Chen, "Probabilistic-based scheduling for runtime goal sequence of agents," in Computer Science and Automation Engineering (CSAE), 2012 IEEE International Conference on, vol. 3. IEEE, 2012, pp. 490–494.

[50] Q. Wu, X. Han, H. Yu, Z. Shen, and C. Miao, "The innovative application of learning companions in virtual singapura," in Proceedings of the 2013 international conference on Autonomous agents and multi-agent systems. International Foundation for Autonomous Agents and, 2013, pp. 1171–1172.

[51] H. Yu, Z. Shen, and C. Leung, "From internet of things to internet of agents," in The 2013 IEEE International Conference on Internet of Things (iThings 2013). IEEE Xplore, 2013, pp. 1054–1057.

[52] ——, "Bringing reputation-awareness into crowdsourcing," in The 9th International Conference on Information, Communications and Signal

[53] ——, "Towards health care service ecosystem management for the elderly," International Journal of Information Technology (IJIT), vol. 19, no. 2, 2013.

[54] J. Ji, H. Yu, B. Li, Z. Shen, and C. Miao, "Learning chinese characters with gestures," International Journal of Information Technology (IJIT), vol. 19, no. 1, 2013.

[55] H. Yu, C. Miao, B. An, Z. Shen, and C. Leung, "Reputation-aware task allocation for human trustees," in The 13th International Conference on Autonomous Agents and Multi-Agent Systems (AAMAS'14). IFAAMAS, 2014, pp. 357–364.

[56] H. Yu, Z. Shen, C. Miao, and C. Leung, "A dynamic method for mixing direct and indirect trust evidence," in The 1st International Workshop on Age-friendly Intelligent Computing -the 2012 World Congress on Computational Intelligence (WCCI'12), 2012.

[57] H. Yu, Z. Shen, Q. Wu, and C. Miao, "Designing socially intelligent virtual companions," in Workshop on Autonomous Social Robots and Virtual Humans at the 25th International Conference on Computer Animation and Social Agents (CASA'12), 2012.

[58] H. Yu, X. Yu, S. F. Lim, J. Lin, Z. Shen, and C. Miao, "A multi-agent game for studying human decision-making," in The 13th International Conference on Autonomous Agents and Multi-Agent Systems (AAMAS'14), 2014, pp. 1661–1662.

[59] Y. Cai, Z. Shen, S. Liu, H. Yu, X. Han, J. Ji, C. Miao, M. J. McKeown, C. Leung, and C. Miao, "An agent-based game for the predictive diagnosis of parkinson's disease," in The 13th International Conference on Autonomous Agents and Multi-Agent Systems (AAMAS'14), 2014, pp. 1663–1664.

[60] Y. Liu, J. Zhang, H. Yu, and C. Miao, "Reputation-aware continuous double auction," in The 28th AAAI Conference on Artificial Intelligence

(AAAI-14). The AAAI Press, 2014.

[61] Y. Liu, S. Liu, H. Fang, J. Zhang, H. Yu, and C. Miao, "Reprev: Mitigating the negative effects of misreported ratings," in The 28th AAAI Conference on Artificial Intelligence (AAAI-14). The AAAI Press, 2014.

[62] J. Lin, H. Yu, Z. Shen, and C. Miao, "Using goal net to model user stories in agile software development," in The 15th IEEE/ACIS International Conference on Software Engineering, Artificial Intelligence, Networking and Parallel/Distributed Computing (SNPD'14), 2014, pp. 1–6.

[63] H. Yu and Y. Tian, "Developing multiplayer mobile game using midp 2.0 game api and jsr-82 java bluetooth api," in The 2005 Cybergames Conference, 2005.

[64] H. Yu, Z. Shen, X. Li, C. Leung, and C. Miao, "Whose opinions to trust more, your own or others'?" The 1st Workshop on Incentives and Trust in E-commerce the 13th ACM Conference on Electronic Commerce (WIT-EC'12), pp. 1–12, 2012.

[65] H. Yu, Z. Shen, C. Miao, B. An, and C. Leung, "Filtering trust opinions through reinforcement learning," Decision Support Systems (DSS), vol. 66, pp. 102–113, 2014.

[66] J. Lin, H. Yu, Z. Shen, and C. Miao, "Studying task allocation decisions of novice agile teams with data from agile project management tools," in The 29th IEEE/ACM International Conference on Automated Software Engineering (ASE'14), 2014, pp. 689–694.

[67] J.-P. Mei, H. Yu, Y. Liu, Z. Shen, and C. Miao, "A social trust model considering trustees' influence," in The 17th International Conference on Principles and Practice of Multi-Agent Systems (PRIMA'14), 2014, pp. 357–364.

[68] H. Yu, C. Miao, Z. Shen, C. Leung, Y. Chen, and Q. Yang, "Efficient task sub-delegation for crowdsourcing," in The 29th AAAI Conference on Artificial Intelligence (AAAI-15). AAAI Press, 2015, pp. 1305–1311.

[69] H. Yu, C. Miao, Z. Shen, and C. Leung, "Quality and budget aware task

allocation for spatial crowdsourcing," in The 14th International Conference on Autonomous Agents and Multi-Agent Systems (AAMAS'15), 2015, pp. 1689–1690.

[70] H. Yu, H. Lin, S. F. Lim, J. Lin, Z. Shen, and C. Miao, "Empirical analysis of reputation-aware task delegation by humans from a multi-agent game," in The 14th International Conference on Autonomous Agents and Multi-Agent Systems (AAMAS'15), 2015, pp. 1687–1688.

[71] J. Lin, H. Yu, C. Miao, and Z. Shen, "An affective agent for studying composite emotions," in The 14th International Conference on Autonomous Agents and Multi-Agent Systems (AAMAS'15), 2015, pp. 1947–1948.

[72] C. Leung, Z. Shen, H. Zhang, Q. Wu, J. C. Leung, K. H. Pang, H. Yu, and C. Miao, "Aging in-place: From unobtrusive sensing to graceful aging," in The 24th Annual John K. Friesen Conference "Harnessing Technology for Aging-in-Place", 2015.

[73] S. Liu, C. Miao, Y. Liu, H. Fang, H. Yu, J. Zhang, and C. Leung, "A reputation revision mechanism to mitigate the negative effects of misreported ratings," in The 17th International Conference on Electronic Commerce (ICEC'15), 2015.

[74] S. Liu, C. Miao, Y. Liu, H. Yu, J. Zhang, and C. Leung, "An incentive mechanism to elicit truthful opinions for crowdsourced multiple choice consensus tasks," in The 2015 IEEE/WIC/ACM International Joint Conference on Web Intelligence and Intelligent Agent Technology (WI-IAT'15), 2015.

[75] B. Li, H. Yu, Z. Shen, L. Cui, and V. R. Lesser, "An evolutionary framework for multi-agent organizations," in The 2015 IEEE/WIC/ACM International Joint Conference on Web Intelligence and Intelligent Agent Technology (WI-IAT'15), 2015.

[76] Z. Pan, C. Miao, H. Yu, C. Leung, and J. J. Chin, "The effects of familiarity design on the adoption of wellness games by the elderly," in

The 2015 IEEE/WIC/ACM International Joint Conference on Web Intelligence and Intelligent Agent Technology (WI-IAT'15), 2015.

[77] Z. Pan, C. Miao, B. T. H. Tan, H. Yu, and C. Leung, "Agent augmented inter-generational crowdsourcing," in The 2015 IEEE/WIC/ACM International Joint Conference on Web Intelligence and Intelligent Agent Technology (WI-IAT'15), 2015.

[78] H. Lin, J. Hou, H. Yu, Z. Shen, and C. Miao, "An agent-based game platform for exercising people's prospective memory," in The 2015 IEEE/WIC/ACM International Joint Conference on Web Intelligence and Intelligent Agent Technology (WI-IAT'15), 2015.

[79] H. Yu, C. Miao, S. Liu, Z. Pan, N. S. B. Khalid, Z. Shen, and C. Leung, "Productive aging through intelligent personalized crowdsourcing," in The 30th AAAI Conference on Artificial Intelligence (AAAI-16), 2016.

[80] Z. Shen, H. Yu, C. Miao, S. Li, and Y. Chen, "Multi-agent system development made easy," in The 30th AAAI Conference on Artificial Intelligence (AAAI-16), 2016.

[81] Y. Shi, C. Sun, Q. Li, L. Cui, H. Yu, and C. Miao, "A fraud resilient medical insurance claim system," in The 30th AAAI Conference on Artificial Intelligence (AAAI-16), 2016.

[82] H. Yu, C. Miao, Z. Shen, J. Lin, and C. Leung, "Infusing human factors into algorithmic crowdsourcing," in The 28th Conference on Innovative Applications of AI (IAAI-16), 2016.

[83] Z. Pan, H. Yu, C. Miao, and C. Leung, "Efficient collaborative crowdsourcing," in The 30th AAAI Conference on Artificial Intelligence (AAAI-16), 2016.

[84] H. Lin, H. Yu, C. Miao, and L. Qiu, "Towards emotionally intelligent machines: Taking social contexts into account," in The 18th International Conference on Human-Computer Interaction (HCI'16), 2016.

[85] Z. Man, K. Lee, D. Wang, Z. Cao, and C. Miao, "A new robust training algorithm for a class of single-hidden layer feedforward neural networks,"

Neurocomputing, vol. 74, no. 16, pp. 2491–2501, 2011.

[86] L. Pan, X. Luo, X. Meng, C. Miao, M. He, and X. Guo, "A two-stage winwin multiattribute negotiation model: Optimization and then concession," Computational Intelligence, vol. 29, no. 4, pp. 577–626, 2013.

[87] Y. Zhao, M. Ma, C. Miao, and T. Nguyen, "An energy-efficient and low-latency mac protocol with adaptive scheduling for multi-hop wireless sensor networks," Computer Communications, vol. 33, no. 12, pp. 1452–1461, 2010.

[88] J. Weng, C. Miao, A. Goh, Z. Shen, and R. Gay, "Trust-based agent community for collaborative recommendation," in Proceedings of the 5th international joint conference on Autonomous agents and multi-agent systems (AAMAS'06), 2006, pp. 1260–1262.

[89] H. Song, C. Miao, Z. Shen, W. Roel, D. Maja, and C. Francky, "Design of fuzzy cognitive maps using neural networks for predicting chaotic time series," Neural Networks, vol. 23, no. 10, pp. 1264–1275, 2010.

[90] H. Song, Z. Shen, C. Miao, Y. Miao, and B. S. Lee, "A fuzzy neural network with fuzzy impact grades," Neurocomputing, vol. 72, no. 13, pp. 3098–3122, 2009.

[91] C. Miao, Q. Yang, H. Fang, and A. Goh, "Fuzzy cognitive agents for personalized recommendation web information systems engineering," in Proceedings of the 3rd International Conference on Web Information Systems Engineering (WISE'02), 2002, pp. 362–371.

[92] Y. Zhao, C. Miao, M. Ma, J. Zhang, and C. Leung, "A survey and projection on medium access control protocols for wireless sensor networks," ACM Computing Surveys, vol. 45, no. 1, 2012.

[93] D. Domazet, C. Miao, C. Calvin, H. Kong, and A. Goh, "An infrastructure for inter-organizational collaborative product development system sciences," in Proceedings of the 33rd Annual Hawaii International Conference on System Sciences, 2000.

[94] G. Zhao, Z. Shen, C. Miao, and Z. Man, "On improving the conditioning

of extreme learning machine: a linear case," in Proceedings of the 7th International Conference on Information, Communications and Signal Processing (ICICS'09), 2009.

[95] C. Miao, A. Goh, Y. Miao, and Z. Yang, "A dynamic inference model for intelligent agents," International Journal of Software Engineering and Knowledge Engineering, vol. 11, no. 5, pp. 509–528, 2001.

# Appendix 1 Title and Abstract of Conference Paper Submitted

## Title

Virtual World based Educational Data Mining

## Abstract

Highly interactive game-like virtual environment has gained increasing spotlight in academic and educational researches. Besides being an efficient and engaging educational tool, virtual environment also has the potential to be integrated with Educational Data Mining (EDM) to cater to emerging requirements of educational assessment. Nowadays, the traditional academic assessment approaches cannot thoroughly reflect the students' learning competencies which are crucial for them to thrive in a fast-changing world. We propose an assessment system that seamlessly integrates EDM with functionality and affordance of a virtual environment to assess students' learning competency through analysing their behavioural data and patterns. We also propose a set of metrics which can be used for judging students' learning competency and how these metrics can be evaluated computationally by quantifying and capturing students' behavioural data in a virtual environment. The field study, which is conducted in Xinmin Secondary School in Singapore, showed that the proposed assessment system is promising in identifying useful behaviour metrics for assessing learning competency. The system also exhibits more potential in terms of its quantitative and objective approach, comparing to traditional assessment methods.

## Keywords

Educational Data Mining; Virtual Environment; Competency Assessment; Self-Directed Learning

# Appendix 2 The Complete Goal Net Model in the Goal Net Designer

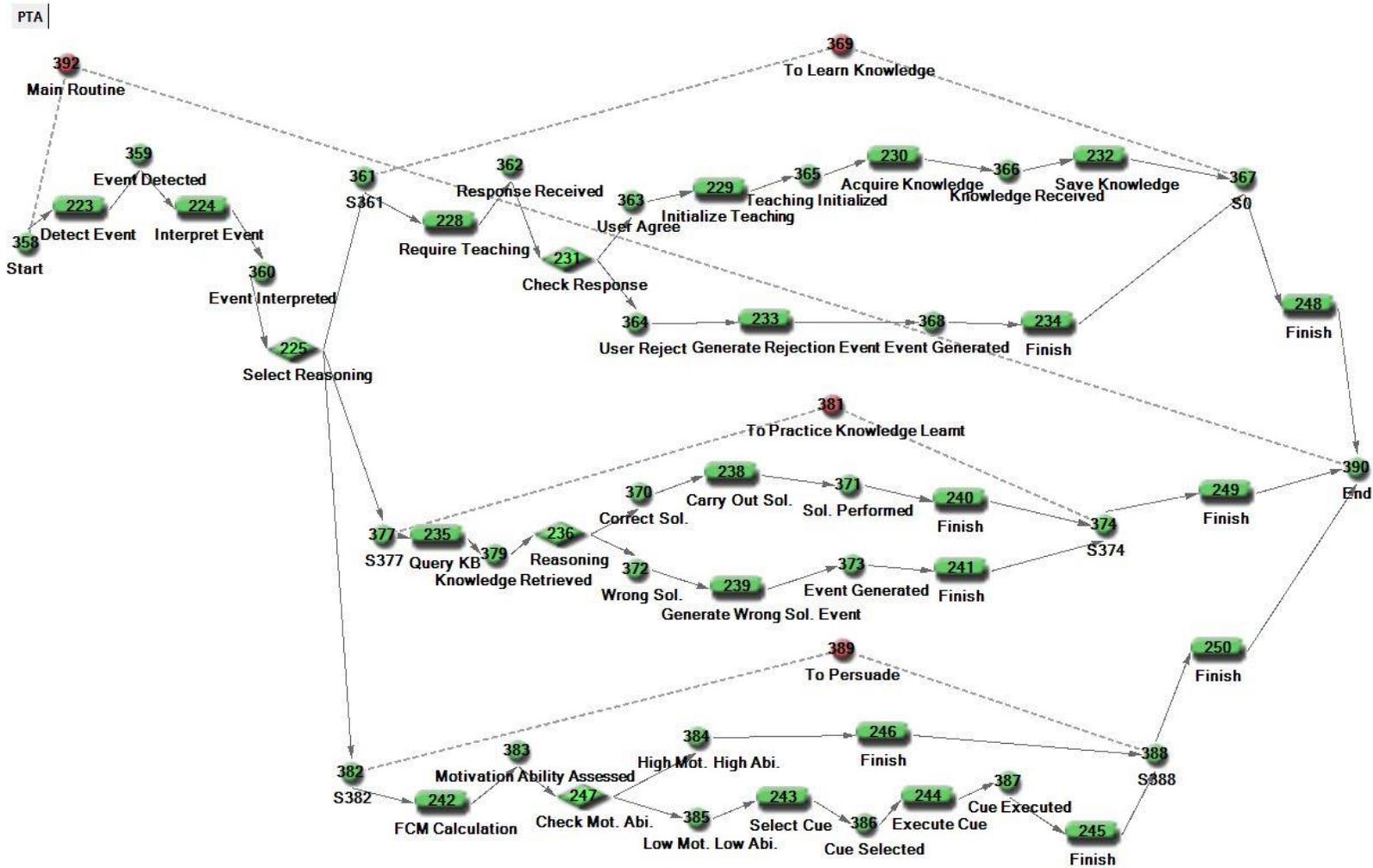

# Appendix 3 Events Tracked in VS Saga

| **Dialogue Events** | |
|---|---|
| Not Learning | Willing to Conduct Experiments |
| Visit Lab | Help Mayor |
| Learn Diffusion | Not Teach Water Molecule |
| Learn Osmosis | Teach Water Molecule |
| Apply Diffusion | Chat with Animal |
| Apply Osmosis | Chat with Village Girl |
| Not Conducting Experiments | Teachability Event |
| **Time Event** | |
| Doing Nothing (Time-Out) | |
| **Teaching Feedback Events** | |
| Teach Success | Teach Failure |
| Practicability Event * | |

* Practicability Event does not belong to any of the three types described above. Both Practicability and Teachability Event are administrative events. However, Teachability Event is generated within the dialogue system of Unity, while Practicability Event is not.

# Appendix 4 The PTA FCM Used in VS Saga

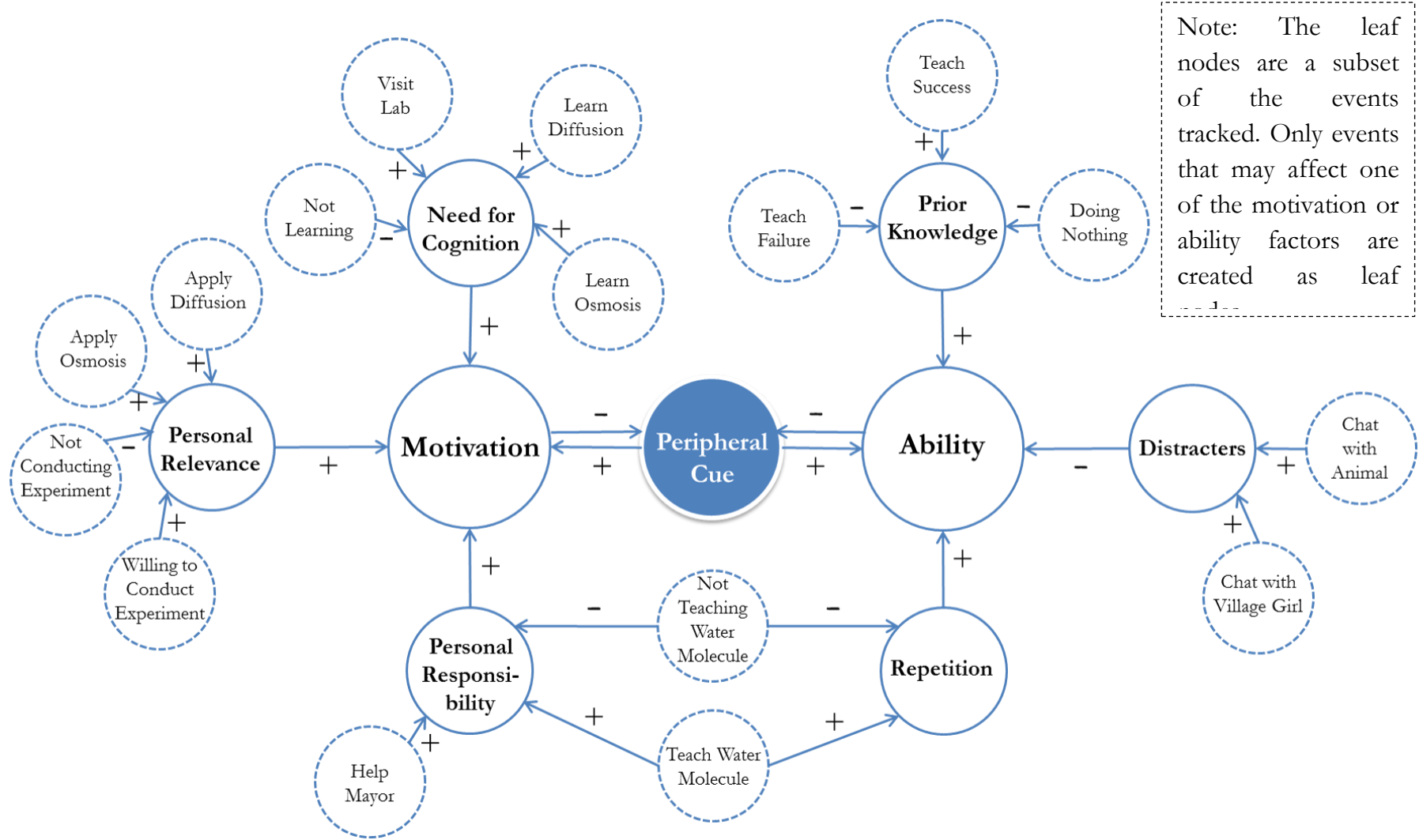

Note: The leaf nodes are a subset of the events tracked. Only events that may affect one of the motivation or ability factors are created as leaf

# Appendix 5 Complete Class Diagram of Agent Control

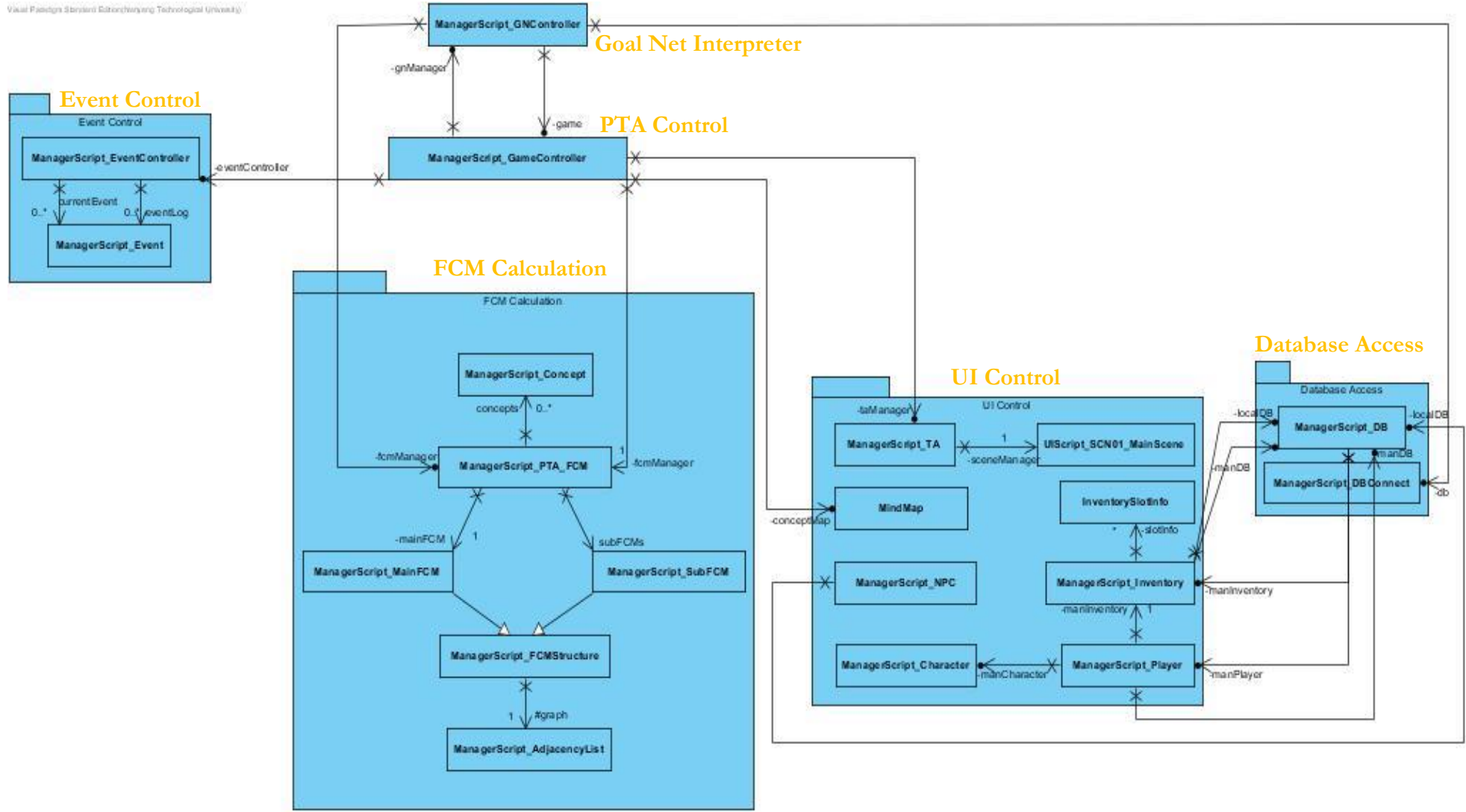